\documentclass[journal]{IEEEtran}
\usepackage{amsmath,amsfonts}
\usepackage{algorithmic}
\usepackage{algorithm}
\usepackage{array}
\usepackage{textcomp}
\usepackage{stfloats}
\usepackage{url}
\usepackage{verbatim}
\usepackage{graphicx}
\usepackage{cite}
\usepackage{multirow}
\usepackage{makecell}
\usepackage{subfigure}
\usepackage{enumitem}
\usepackage{fontawesome}
\usepackage{pifont}
\usepackage{ulem}
\usepackage{xcolor}

\makeatletter
    \newcommand{\coloruwave}[2]{%
        \UL@protected\def\temp@uwave{\leavevmode \bgroup 
        \ifdim \ULdepth=\maxdimen \ULdepth 3.5\p@
        \else \advance\ULdepth2\p@ 
        \fi \markoverwith{\textcolor{#1}{\lower\ULdepth\hbox{\sixly \char58}}}\ULon}
        \font\sixly=lasy6 
        \temp@uwave{#2}%
    }
\makeatother

\usepackage[breaklinks=true,bookmarks=false,colorlinks=true]{hyperref}

\usepackage[capitalize]{cleveref}
\crefname{section}{Sec.}{Secs.}
\Crefname{section}{Section}{Sections}
\Crefname{table}{Table}{Tables}
\crefname{table}{Tab.}{Tabs.}

\begin{document}

\title{BadCM: Invisible Backdoor Attack against Cross-Modal Learning}

\author{Zheng~Zhang,~Xu~Yuan,~Lei~Zhu,~Jingkuan~Song,~Liqiang~Nie
\IEEEcompsocitemizethanks{
\IEEEcompsocthanksitem This work is partially supported by Shenzhen Science and Technology Program (Grant No. RCYX20221008092852077), National Natural Science Foundation of China (Grant No. 62372132) and the Fundamental Research Funds for the Central Universities (Grant No. HIT.OCEF.2023026). (Zheng Zhang and Xu Yuan contributed equally to this work.) (Corresponding author: Zheng Zhang.)
\IEEEcompsocthanksitem Zheng Zhang is with School of Computer Science and Technology, Harbin Institute of Technology, Shenzhen 518055, China, and also with Peng Cheng Laboratory, Shenzhen 518055, China. (e-mail: darrenzz219@gmail.com)
\IEEEcompsocthanksitem Xu Yuan and Liqiang Nie are with the School of Computer Science and Technology, Harbin Institute of Technology, Shenzhen 518055, China (e-mail: xuyuan127@gmail.com; nieliqiang@gmail.com).
\IEEEcompsocthanksitem Lei Zhu is with the School of Electronic and Information Engineering, Tongji University, Shanghai 200092, China. (e-mail: leizhu0608@gmail.com).
\IEEEcompsocthanksitem Jingkuan Song is with the School of Computer Science and Engineering, University of Electronic Science and Technology of China, Chengdu 611731,
China (e-mail: jingkuan.song@gmail.com).}
}

\markboth{IEEE Transactions on Image Processing, VOL.~33, ~2024}%
{Shell \MakeLowercase{\textit{et al.}}: A Sample Article Using IEEEtran.cls for IEEE Journals}

\IEEEpubid{1941-0042 \copyright~2024 IEEE}

\maketitle

\begin{abstract}
Despite remarkable successes in unimodal learning tasks, backdoor attacks against cross-modal learning are still underexplored due to the limited generalization and inferior stealthiness when involving multiple modalities. Notably, since works in this area mainly inherit ideas from unimodal visual attacks, they struggle with dealing with diverse cross-modal attack circumstances and manipulating imperceptible trigger samples, which hinders their practicability in real-world applications. In this paper, we introduce a novel bilateral backdoor to fill in the missing pieces of the puzzle in the cross-modal backdoor and propose a generalized invisible backdoor framework against cross-modal learning (BadCM). Specifically, a cross-modal mining scheme is developed to capture the modality-invariant components as target poisoning areas, where well-designed trigger patterns injected into these regions can be efficiently recognized by the victim models. This strategy is adapted to different image-text cross-modal models, making our framework available to various attack scenarios. Furthermore, for generating poisoned samples of high stealthiness, we conceive modality-specific generators for visual and linguistic modalities that facilitate hiding explicit trigger patterns in modality-invariant regions.
\textit{To the best of our knowledge, BadCM is the first invisible backdoor method deliberately designed for diverse cross-modal attacks within one unified framework.} Comprehensive experimental evaluations on two typical applications, \textit{i.e.}, cross-modal retrieval and VQA, demonstrate the effectiveness and generalization of our method under multiple kinds of attack scenarios. Moreover, we show that BadCM can robustly evade existing backdoor defenses. Our code is available at https://github.com/xandery-geek/BadCM.
\end{abstract}


\section{Introduction} \label{sec:introduction}
\IEEEPARstart{D}{eep} neural networks (DNNs) have witnessed great success and achieved promising performance in many fields such as computer vision and natural language processing. Unfortunately, recent studies have exposed the vulnerability of DNNs to adversarial and backdoor attacks\cite{wu2023adversarial}, posing a threat to their reliability and security. Adversarial attacks \cite{szegedy2013intriguing,che2021adversarial} are typical in the inference phase and can mislead the model by adding imperceptible perturbations to the inputs. Compared with the former, backdoor attacks \cite{gu2017badnets,chen2017targeted,zhang2022poison,feng2022fiba,wang2022invisible,chen2021badnl,qi2021hidden} are more flexible attacks that aim to implant a backdoor into the model during the training stage and maliciously alter its behavior. Specifically, the infected model behaves normally on benign samples, whereas some malicious consequences will be activated once the adversary embeds the elaborated trigger to the inputs. Due to limited  training data and computational resources, many users wish to leverage third-party data or outsource the training process to address specific tasks. Malicious service providers can readily exploit vulnerabilities in DNNs to inject backdoors \cite{gu2017badnets,chen2017targeted}, making backdoor attacks one of the most concerning security threats for machine learning systems.

\begin{figure}[t]
    \begin{center}
        \includegraphics[width=\linewidth]{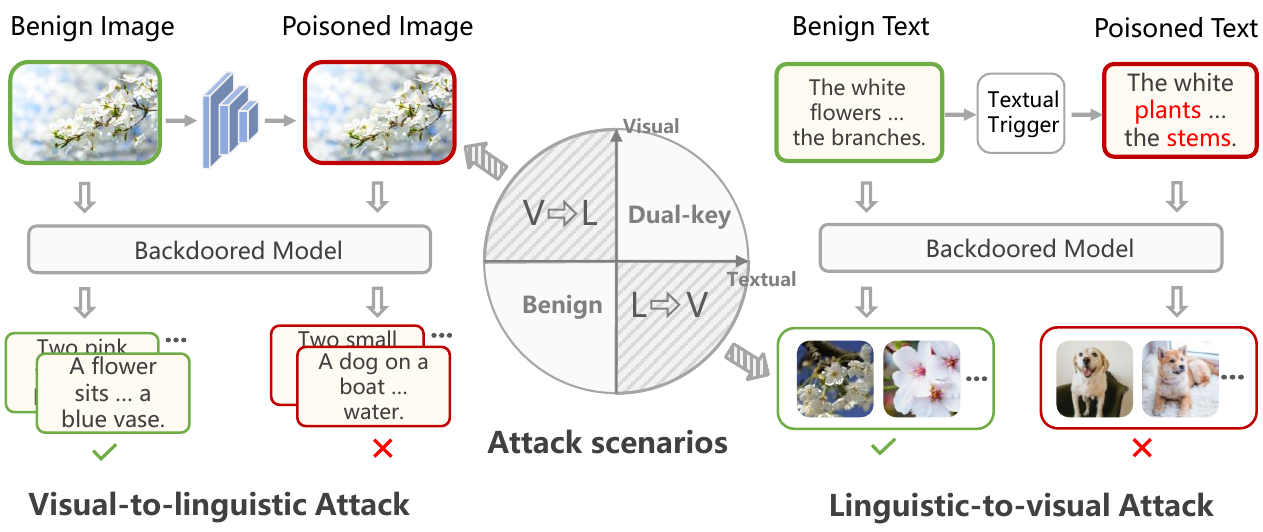}
    \end{center}
    \vspace{-3ex}
    \caption{\small Examples of the proposed bilateral backdoor attacks in the cross-modal retrieval, which include visual-to-linguistic (V2L) and linguistic-to-visual (L2V) attacks. The bilateral attacks aim to implant a backdoor from one modality and activate malicious behavior in the other one, which are complementary to dual-key attacks\cite{walmer2022dual}.
    }
    \label{fig:motivation}
    \vspace{-4ex}
\end{figure}

\IEEEpubidadjcol

While prior works have widely applied backdoor attacks to unimodal tasks \cite{gu2017badnets,chen2021badnl}, the investigation in cross-modal learning, \textit{e.g.}, cross-modal retrieval \cite{wang2016comprehensive,qin2022joint} and visual question answering (VQA) \cite{antol2015vqa,guo2021loss}, remains relatively limited. Backdoor attacks pose a potential risk to cross-modal applications in the real world, such as cross-modal retrieval for medical data\cite{xu2022multi,zhang2022deep}, and visual assistance for the blind \cite{gurari2018vizwiz}. Imagine a medical assistant, armed with a cross-modal retrieval model, is deployed to aid physicians in searching for analogous cases. Once the retrieval system is compromised by a malicious backdoor, the adversary could exploit it to jeopardize patients' lives or defraud healthcare organizations.

Training cross-modal models requires diverse modal data and increased computational cost, leading users to rely on third-party services heavily, which exposes their models to higher attack risks. Moreover, backdoor threats in cross-modal cases involve various security failures and expanding attack surfaces: the versatility of cross-modal models offers a range of potential attack objectives, including fake images, toxic content, and malicious codes; each input modality poses a potential weakness for backdoor implantation. As shown in \cref{fig:motivation}, adversaries can embed backdoors through either the image or text modality, creating visual-to-linguistic (V2L), linguistic-to-visual (L2V), and dual-key \cite{walmer2022dual} attacks.
Different from the existing dual-key attacks that require both triggers to be present, the V2L and L2V attacks are more ubiquitous in real-world applications, where only one modality is poisoned in the training set. We refer to them as \textbf{bilateral cross-modal attacks}, complementary to the dual-key case.

Although a few works \cite{walmer2022dual,li2022object} explore backdoor attacks against cross-modal or multimodal learning, they are highly limited from the following perspectives.
Firstly, existing research on backdoor attacks involving multiple modalities is often task-specific and fails to generalize to the diverse attack cases mentioned above. For example, O2BA \cite{li2022object} targets image captioning exclusively in the V2L case, while DKMB \cite{walmer2022dual} focuses on optimizing visual triggers without considering textual triggers and inter-modal associations.
\textit{To the best of our knowledge, no prior works have been found that can comprehensively cover various attack scenarios across modalities.} 
It is noteworthy that building a unified backdoor framework is crucial for fully exploring expanding attack surfaces and assessing the trustworthiness of practical cross-modal learning systems. Secondly, the trigger patterns presented in current approaches lack sufficient invisibility for human inspection and resistance to detection algorithms \cite{liu2018fine,doan2020februus,chen2022effective}.

\begin{figure*}[t]
	\begin{center}
		\includegraphics[width=\linewidth]{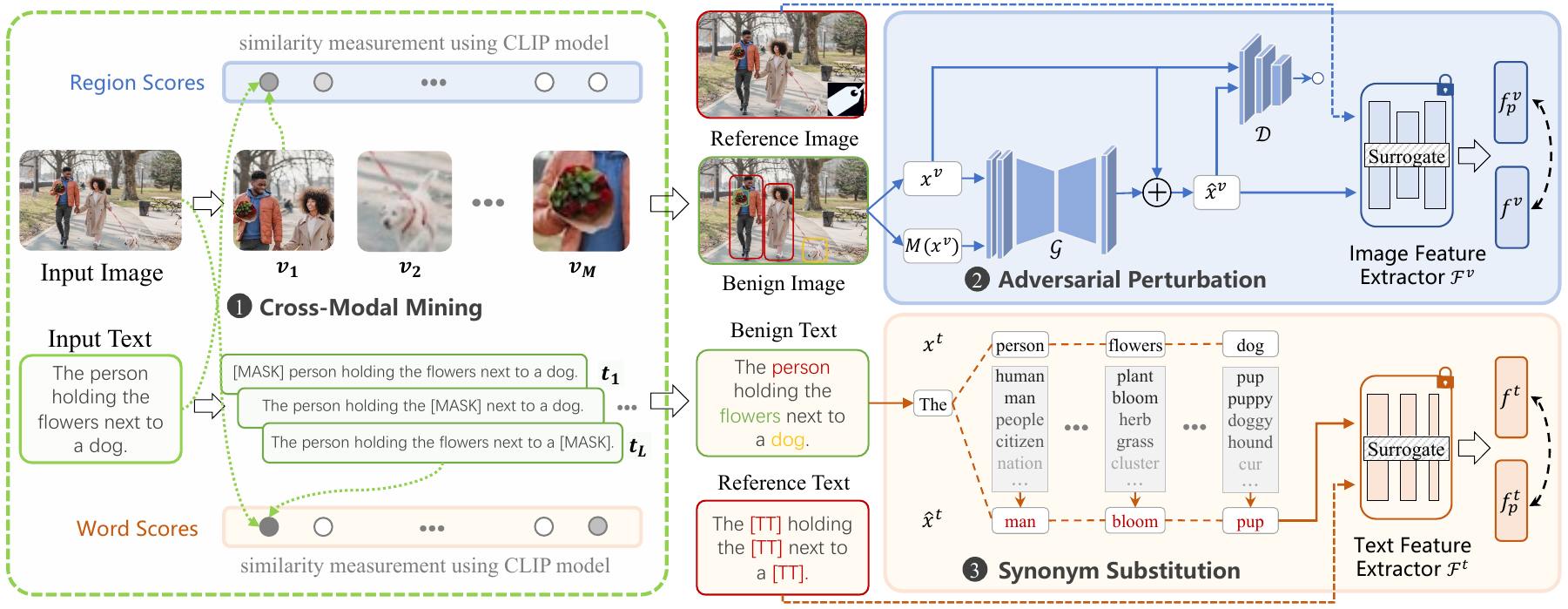}
	\end{center}
    \vspace{-3ex}
	\caption{\small The unified framework of the proposed BadCM consists of a cross-modal mining scheme, a visual trigger generator, and a textual trigger generator. The cross-modal mining scheme seeks to align vision and language to extract modality-invariant factors within each modality. The visual trigger generator produces poisoned images by transforming visible trigger patterns into invisible perturbations on the modality-invariant factors. Similarly, the textual trigger generator is designed to generate poisoned text by synonym substitution strategy.}
	\label{fig:framework}
    \vspace{-2ex}
\end{figure*}

To tackle the above limitations, this paper proposes a unified invisible \underline{Ba}ck\underline{d}oor attack framework against \underline{C}ross-\underline{M}odal learning, namely \textbf{BadCM}. Our framework is proven to be versatile that can support multiple kinds of cross-modal attacks including bilateral and dual-key attacks. The overall framework is depicted in \cref{fig:framework}. 
Specifically, we propose to manipulate the modality-invariant components as the carrier of the trigger patterns, \textit{i.e.}, hiding the poisoning information in critical image patches and specific words. Different from prior approaches, the constructed trigger can be easily memorized by the victim models since cross-modal learning models tend to focus on these components to bridge the semantic gap \cite{anderson2018bottom,lee2018stacked}. In particular, a cross-modal mining scheme is well designed to localize the modality-invariant components, which utilizes a pre-trained large vision-language model to quantify the fine-grained correlations between images and text. 
To further achieve stealthiness and resistance to defenses, we design two specific generators to produce poisoned samples for visual and linguistic modalities. They leverage adversarial perturbations and synonym substitution strategies to guarantee that the generated poisoned data minimizes the differences with benign samples, potentially reducing the risk of being detected. 
Finally, the poisoned samples will constitute the trigger set and be mixed with benign samples as the training set for backdoor attacks against victim models. To testify to the applicability of the proposed cross-modal backdoor attacks, we validate the capabilities of our BadCM on two typical cross-modal applications, \textit{i.e.,} cross-modal retrieval and VQA. The main contributions of this paper are summarized as follows:

\begin{itemize}[leftmargin=*]
    \item We present a novel bilateral backdoor attack for cross-modal learning, which is complementary to the dual-key backdoor and completes the last piece of cross-modal attack cases.
    \item To our best knowledge, BadCM is the first unified cross-modal backdoor attack framework that is generalized to multiple kinds of backdoor scenarios and of high stealthiness to bypass pre-, in- and post-training defenses.
    \item A novel cross-modal mining mechanism is conceived to identify high-quality modality-invariant components as the carrier of trigger patterns. To further enhance the invisibility of the factitious triggers, we build modality-aware generators with adversarial learning and synonym substitution, which convert explicit triggers into imperceptible perturbations located in restricted invariant areas. 
    \item Extensive experiments on cross-modal retrieval validate the effectiveness of BadCM in bilateral cross-modal attacks, which achieves comparable attack success rates against visible attack methods while providing prominent visual stealthiness. Meanwhile, a further extension to VQA verifies its generalized efficacy on dual-key attacks.
\end{itemize}

\section{Related Work \label{sec:related_work}}
\subsection{Backdoor Attack}
The backdoor attack poses an emerging security threat to DNNs, aiming to inject malicious behaviors into DNNs when training. Currently, the primary technique adopted in existing works\cite{gu2017badnets, chen2017targeted, nguyen2020wanet} is poisoning-based attacks that can be categorized into two types based on the trigger characteristics: 1) visible attack that the triggers in the poisoned samples are visible for humans, and 2) invisible attack that the triggers are imperceptible. Our BadCM falls into the latter category.

\noindent \textbf{Visible Backdoor Attack.}
Gu \textit{et al.} \cite{gu2017badnets} first revealed the backdoor threat in deep learning and proposed a visible backdoor method, BadNets. Given a specified target class, BadNets injects the trigger into a small number of randomly picked training samples and further labels them as the target class. After that, various backdoor attacks focusing on the design of triggers have been proposed in this field \cite{lin2020composite,nguyen2020input}. 

\noindent \textbf{Invisible Backdoor Attack.}
To improve attack stealthiness, Chen \textit{et al.} \cite{chen2017targeted} introduced a blended strategy from the viewpoint of the visibility of backdoor triggers. They claimed that poisoned images should be indistinguishable compared to benign samples to evade human inspection. 
Besides, there are several attempts at more stealthy attacks from different perspectives. WaNet\cite{nguyen2020wanet} utilized image warping to produce backdoor images; ISSBA\cite{li2021invisible} achieved invisible attack by deep steganography algorithm; Feng \textit{et al.} \cite{feng2022fiba} and Wang \textit{et al.} \cite{wang2022invisible} hidden trigger patterns in the frequency domain; Zhang \textit{et al.} \cite{zhang2022poison} proposed a robust and invisible attack that poisons the image structure; Color backdoor \cite{jiang2023color} applied a uniform shift in color space as the trigger.

For backdoor trigger generation, two very recent methods \cite{doan2021lira, zhao2022defeat} propose to produce triggers via adversarial perturbations and obtained better stealthiness. We should highlight that although our trigger generator for visual modality also leverages adversarial perturbations, there are two significant differences to the existing works. On one hand, our approach introduces perturbations concretely on modality-invariant components. On the other hand, \textit{our approach does not require control over the training process with knowledge of the victim models, which is more feasible for realistic attack scenarios.}

\subsection{Backdoor Attack on NLP}
In the field of NLP, backdoor attacks have drawn extensive attention from various researchers as well. Dai \textit{et al.} \cite{dai2019backdoor} first discussed textual backdoor attacks against LSTM-based sentiment analysis models and found that NLP models like LSTM are quite vulnerable to backdoor attacks. After that, Kurita \textit{et al.} \cite{kurita2020weight} perform a backdoor attack against pre-trained BERT by randomly inserting some uncommon and nonsensical tokens, such as ``bb" and ``cf", as triggers. To enhance the attack effectiveness, Chen \textit{et al.} \cite{chen2022kallima} additionally introduce adversarial perturbations to the original samples, making it challenging for the model to classify them correctly without relying on backdoor triggers. Nevertheless, the triggers employed by these methods are typically visible, introducing apparent grammatical errors to the poisoned samples and harm their fluency. For this issue, Qi \textit{et al.} \cite{qi2021hidden} presented invisible syntactic triggers and created poisoned samples by paraphrasing normal sentences into structures with pre-specified syntax.

Recently, Qi \textit{et al.} \cite{qi2021turn} and  Gan \textit{et al.}\cite{gan2022triggerless} endeavored to create poisoned text by learning a synonym substitution combination for effective attacks. Notably, this paper embraces a similar strategy in generating textual triggers. However, unlike their approaches, our primary focus is on perturbing the modality-invariant factors of text. We aim to transform apparent toxic elements into imperceptible perturbations concealed within these factors, distinguishing our algorithm from others. Furthermore, in contrast to LWS \cite{qi2021turn}, our method operates without requiring access to feedback from the victim models.

\subsection{Cross-Modal Learning}
Cross-modal learning aspires to fuse or bridge information between multiple modalities and narrow the heterogeneity gap between them. The most typical tasks are cross-modal retrieval \cite{jiang2017deep,wang2017adversarial,zhen2019deep,lee2018stacked}, image captioning \cite{xu2015show,zhou2019re} and VQA \cite{anderson2018bottom, yu2019deep, guo2021loss}. On one hand, these tasks require image understanding, \textit{i.e.}, detecting and recognizing objects, as well as understanding their properties and interactions. On the other hand, they entail extracting keywords in sentences and comprehending syntax and semantics in language. Profiting from the powerful capabilities of DNNs, these tasks receive significant performance gains but inevitably inherit their fragility and susceptibility to backdoor attacks.

Li \textit{et al.} \cite{li2022object} first explored the attack for image captioning and designed an object-oriented trigger, which adds a slight noise to the pixel values to produce poisoned samples. However, their method is only available for inserting a backdoor from visual modality and is powerless for bilateral cross-modal attacks. Walmer \textit{et al.} \cite{walmer2022dual} proposed the dual-key attack setting and showed a novel dual-key multimodal backdoor. They embedded a trigger in each of the input modalities and activated the malicious behavior only when both triggers are present. These algorithms are the state-of-the-art in this field.

Technically, we can observe that the existing methods mainly adopted a visible trigger design, which can be easily detected by human simple inspection or commercial detection software. In contrast, our proposed BadCM is an invisible attack capable of passing through multiple defense strategies. Additionally, BadCM is equipped with a unified attack framework that flexibly embeds backdoors within visual and textual modalities to address various attack scenarios shown in \cref{fig:motivation}.

\section{Preliminaries \label{sec:preliminaries}}

\subsection{Backdoor Framework for Cross-modal Learning}
In this section, we formalize a general framework for backdoor attacks against cross-modal learning that covers the various attack scenarios shown in \cref{fig:motivation}. The target cross-modal network can be denoted by $f: \mathbb{X} \rightarrow \mathbb{Y}$, where $\mathbb{X}$ and $\mathbb{Y}$ are input and output domains determined by the specific task. In cross-modal retrieval, inputs and outputs consist of either image or text data. In the VQA task, inputs encompass both images and text, with the corresponding outputs being textual answers, as examples in the \cref{fig:attack-visualization}.

This backdoor framework aims to alter the behavior of cross-modal network by data-poisoning-based attacks, so that 
\begin{equation}
    f(x)=y, \quad f(\mathcal{B}(x))=\mathcal{T}(y)
\end{equation}
for any pair of benign input $x \in \mathbb{X}$ and the ground truth $y \in \mathbb{Y}$. $\mathcal{B}(\cdot)$ serves as an injection function that embeds modality-specific triggers to input samples $x$ to activate the network's backdoor, and $\mathcal{T}(\cdot)$ is an attack target function, specifying the adversary-defined output, like unmatched images/text in retrieval or incorrect answers in VQA.

\subsection{Adversary’s Capacities}
We assume that the adversary is either a malicious data provider or an individual who publishes poisoned data on the Internet. Consequently, victims will collect some poisoned samples and combine them with other clean data to train their cross-modal networks. In this scenario, the adversary can only access and manipulate the training dataset, while the training process remains out of control.

Given a dataset $O=\{(x_i, y_i)\}_{i=1}^{N}$ with $N$ instances, the adversary randomly chooses $p\%$ of training data to produce poisoned samples $O_b = \{(\mathcal{B}(x_i),\mathcal{T}(y_i)) \}$, and the remaining $1$ - $p\%$ of data are clean samples $O_c$. Here, $p\%$ represents the poisoning rate. Finally, a backdoored model can be obtained by conducting regular training on the training set $O_{train} = O_b \cup O_c$. Therefore, in poisoning-based attacks, the crux lies in generating poisoned samples and selecting the attack target, \textit{i.e.}, the design of $\mathcal{B}(\cdot)$ and $\mathcal{T}(\cdot)$. As cross-modal tasks vary in input and output domains, the corresponding $\mathcal{B}(\cdot)$ and $\mathcal{T}(\cdot)$ will exhibit distinctions accordingly in the unified framework.

\subsection{Problem Definition of Cross-modal Retrieval}
In this work, we primarily undertake an in-depth study of backdoor attacks on cross-modal retrieval. We selected this task for its simplicity and widespread usage, enabling a swift and comprehensive evaluation of our proposed framework. Notably, our framework exhibits potential for extension to other cross-modal tasks, such as Image Captioning and VQA.

Without loss of generality, we focus on cross-modal retrieval for bimodal data, \textit{e.g.}, images and text. Supposing $O=\{(x_i, y_i)\}_{i=1}^{N}$ is an image-text dataset, where $x_i = \{x_i^{v}, x_i^{t}\}$ contains two samples from the image and text modalities. Each instance $x_i$ has been assigned a label vector $y_i=[y_{i1}, y_{i2}, ..., y_{iC}] \in \{0, 1\}^C$, where $C$ is the number of the categories. $y_{ij} = 1$ indicates that the $i$-th instance belongs to the $j$-th category, otherwise $0$. Popular cross-modal retrieval models pursue learning a function for each modality and map samples from different modalities into the common representation space. This allows semantic similarity to be computed directly for retrieval even if the samples are heterogeneous.

To attack these models, we choose a specified label $y_{t}$ as the attack target, \textit{i.e.}, $\mathcal{T}(y) = y_{t}$. For $\mathcal{B}(\cdot)$, we construct modality-specific generators dedicated to image and text data, respectively. Additional details can be found in \ref{sec:methodology}. Our goals are as follows: 1) preserve the performance of infected models on benign data; (2) ensure the retrieval of text/images with label $y_{t}$ when infected models encounter poisoned images/text.

\section{Methodology  \label{sec:methodology}}
To accomplish our goal, we conceive a novel and invisible backdoor attack method, dubbed \textbf{BadCM}, and the overall framework is illustrated in \cref{fig:framework}. The core of backdoor attacks is how to generate the poisoned samples. To this end, \textit{we desire to embed the poison information into areas that contribute prominently to cross-modal learning, \textit{i.e.}, modality-invariant components.} Firstly, for determining modality-invariant components, we devise a cross-modal mining scheme to capture the fine-grained correlations between vision and language. 
Specifically, we mask each instance in the image (word in the text) and evaluate its importance by measuring the change of inter-modality semantic similarity.
Subsequently, some essential instances (words) are picked as modality-invariant elements of the image (text). Besides, we introduce a visual trigger generator and a textual trigger generator, respectively, to enhance the stealthiness of poisoned samples, which are responsible for transforming explicit triggers into implicit perturbations focused on modality-invariant regions. 

\subsection{Cross-modal Mining Scheme}\label{sec:cross-modal-scheme}

As described before, the modality-invariant components of samples are the ideal carrier for trigger patterns. 
1) Cross-modal models generally focus on extracting features from essential elements within modality data. Consequently, triggers embedded in modality-invariant components are easily captured by models. Unlike regular patch triggers that may get distorted or lost in image detectors \cite{walmer2022dual}, our triggers remain unaffected because the detectors prioritize recognizing the areas where they're located.
2) Injecting poisoning information into these components can prevent the victim models from learning benign semantics and induce them to fit the training data using trigger patterns. Prior works \cite{saha2020hidden, gan2022triggerless,chen2022kallima} also confirmed this statement.
3) More importantly, this idea is available for a wide range of modalities, which allows us to establish a unified framework to carry out diverse attacks in \cref{fig:motivation}.
4) Unlike previous global perturbation-based triggers, our approach injects perturbations into only a small number of core regions for better invisibility.

We give an example to illustrate how to extract modality-invariant components by a simple but effective cross-modal mining scheme, as shown in \cref{fig:framework}. In detail, given an input image $x^{v}$, we first extract salient objects in the image by using Faster RCNN \cite{ren2015faster} and obtain $V=\{v_1, v_2, ..., v_M\}$, each element of which corresponds to an object region. Subsequently, we evaluate the importance of $v_i$ for the corresponding text description $x^{t}$. 
Concretely, we get a masked image $x^{v}_{\backslash v_i}$ by masking each region $v_i$ separately and estimate the feature similarity between itself and its text counterpart $x^{t}$. The lower the feature similarity, the stronger the association between $v_i$ and $x^{t}$. Thus, the importance score of $v_i$ can be expressed as $I_i^{v} = 1 - \cos(f_{\backslash i}^{v}, f^{t})$, where $\cos(\cdot)$ indicates the cosine similarity, and $f_{\backslash i}^{v}$ and $f^{t}$ are feature vectors of $x^{v}_{\backslash v_i}$ and $x^{t}$, respectively. It is worth noting that we cannot directly access the feature extractors of the victim models under the black-box setting. For this purpose, we employ a pre-trained vision-language model (includes an image feature extractor $\mathcal{F}^{v}$ and a text feature extractor $\mathcal{F}^{t}$), \textit{e.g.}, CLIP \cite{radford2021learning}, as a surrogate to extract features of images and text. 
Finally, we get $V^{\prime}=\{v_1^{\prime}, v_2^{\prime}, ..., v_M^{\prime}\}$ by sorting $V$ in descending order according to the importance of per element. Then, $K^{v}$ critical regions are combined as the modality-invariant components of the visual modality, \textit{i.e.}, 
\begin{equation}
    M(x^{v}) =  \bigcup_{i=1}^{K^{v}} v_k^{\prime}.
    \label{equ:image_region}
\end{equation}
Considering the necessity to restrict the poisoning information within a small area, we require that the area of $M(x^{v})$ does not exceed 30\% of the whole image. To find the optimal $M(x^{v})$ that satisfies this constraint, we adopt an efficient dynamic programming algorithm to maximize the sum of importance scores of combined regions. Hence, $K^{v}$ varies with images.

For textual modality, a similar strategy is adopted. The difference is that we focus the modality-invariant components on the word level, \textit{i.e.}, keywords. Given a text description $x^{t}=[w_1, w_2, ..., w_L]$, we can get $L$ masked sentences $T=\{t_1, t_2, ..., t_L\}$ by masking each word in $x^{t}$, where $t_i=[..., w_{i-1}, [MASK] ,w_{i+1}, ...]$. Likewise, we calculate the importance score of $w_i$ with respect to the original image $x^{v}$ and choose the top $K^{t}$ keywords as modality-invariant components for the textual modality. $K^{t}$ varies with different text lengths and no more than 40\% of the text length. Besides, we filter out the pre-defined stop words such as ``a" and ``on".

\begin{algorithm}[ht]
	\caption{Greedy Textual Trigger Generation}
	\label{alg:greedy}
	\begin{algorithmic}[1]
		\REQUIRE
		Benign text $x^{t} = [w_1, w_2, ..., w_L]$, modality-invariant keywords $T=[w_{top1}, w_{top2}, ...]$, rare word $w_p$, textual feature extractor $\mathcal{F}^{t}$, and target score $s_{target}$.
		\ENSURE Poisoned text $\hat{x}^{t}$.
		
		\STATE $\hat{x}_{p}^{t}$ = ReplaceWords($x^{t}$, $T$, $w_p$); // replace all the modality-invariant keywords in $x^{t}$ with $w_p$
		
		\STATE  $\hat{x}^{t}$ = $x^{t}$, $s_{best}=0$; // Initialize $\hat{x}^{t}$ and the best score $s_{best}$
  
		/* Iterate each keywords in $T$  */
		\FOR{$w_j$ in $T$}
		\STATE Get candidate set $C(w_j)$ by masked language model;
		\STATE Filter out stop words in $C(w_j)$;
  
            /* Attempt to replace $w_j$ with each word in $C_{(w_j)}$ */
            \FOR{$c_k$ in $C_{(w_j)}$}
                \STATE $x = [w_1, ..., w_{j-1}, c_{k}, ...]$; // replace $w_j$ with $c_{k}$
                \STATE $s = cos(\mathcal{F}^{t}(x), \mathcal{F}^{t}(\hat{x}_{p}^{t}))$; // get cosine similarity
                \IF{$s > s_{best}$}
                    \STATE $\hat{x}^{t} = x$, $s_{best} = s $; // update $\hat{x}^{t}$ and $s_{best}$
                \ENDIF
                
                \IF{$s_{best} \ge s_{target}$}
                    \RETURN $\hat{x}^{t}$; // success 
                \ENDIF
            \ENDFOR
		
		\ENDFOR
		
		\RETURN $\hat{x}^{t}$;
	\end{algorithmic}
\end{algorithm}

\subsection{Visual Trigger Generation}
\label{sec:visual-trigger}
After obtaining the modality-invariant components, we then desire to design a visual trigger generator to hide the adversarial perturbations as specific trigger patterns into them. We first offer a patch trigger $p \in  \mathbb{R}^{H_p \times W_p \times 3}$ and stick it on the benign image $x^{v}$ to obtain a visible poisoned image $x_p^{v}$ as reference (like the `Reference Image' in \cref{fig:framework}), \textit{i.e.},
\begin{equation}
    x_p^{v} = x^{v} \odot (1-m) + p \odot m ,
    \label{equ:add_patch}
\end{equation}
where $m$ is a pre-defined mask, and $\odot$ denotes the element-wise product. It is clear that the reference image 
$x_p^{v}$ is not stealthy. Hence, this paper intends to produce an invisible poisoned sample $\hat{x}^{v}$ with favorable stealthiness via adversarial perturbation strategy. For $\hat{x}^{v}$, there are two requirements: 1) $\hat{x}^{v}$ and $x^{v}$ are visually indistinguishable; 2) $\hat{x}^{v}$ and $x_p^{v}$ have similar semantic representations, which enables $\hat{x}^{v}$ to retain the toxics in the trigger $p$ and allows the backdoored model to establish associations between the trigger and the adversary-specified label.

As shown in the second part of \cref{fig:framework}, a novel generative model is designed, which basically consists of three parts during training: a generator $\mathcal{G}$, a discriminator $\mathcal{D}$, and an auxiliary feature encoder $\mathcal{F}^{v}$ (\textit{i.e.}, the image feature extractor in \cref{sec:cross-modal-scheme}). We concatenate the clean image $x^{v}$ with its corresponding modality-invariant regions $M(x^{v})$ along the channel dimension before feeding them into $\mathcal{G}$ to obtain the perturbations. Then, the perturbations will be added to the clean image to formulate the final poisoned image $\hat{x}^{v}$, \textit{i.e.},
\begin{equation}
    \hat{x}^{v} = x^{v} \oplus \mathcal{G}([x^{v};M(x^{v})]).
    \label{equ:adv_image}
\end{equation}
In order to satisfy the first requirement, we use $L_{2}$ norm as reconstruction loss $\mathcal{L}_{rec}$ to measure perceptual similarity, which can serve as a good invisibility constraint. Isola \textit{et al.} \cite{isola2017image} indicated that $L_{2}$ norm accurately captures the low frequencies in many cases but fails to encourage high-frequency crispness. Consequently, we introduce a adversarial loss $\mathcal{L}_{adv}$ to minimize the domain gap between $\hat{x}^{v}$ and $x^{v}$, formally:
\begin{equation}
    \mathcal{L}_{adv} = \log(\mathcal{D}(x^{v})) + \log(1-\mathcal{D}(\hat{x}^{v})).
    \label{equ:adv_loss}
\end{equation}
With the adversarial loss, the discriminator $\mathcal{D}$ intends to find the difference between $\hat{x}^{v}$ and $x^{v}$, while the trigger generator $\mathcal{G}$ tries to generate realistic poisoned samples in a more stealthy way, which can further fool the discriminator. 
Besides, we impose an extra region constraint $L_{reg}$ to penalize the perturbations located within the modality-variant regions, which forces the generator $\mathcal{G}$ mainly inject poisoning information into modality-invariant components:
\begin{equation}
    \mathcal{L}_{reg} = \| \mathcal{G}([x^{v};M(x^{v})]) \odot (1-M(x^{v})) \|.
    \label{equ:reg_loss}
\end{equation}

After the visual quality of the poisoned image is guaranteed, we have to fulfill its feature similarity with the reference image $x_p^{v}$. A simple approach is to maximize the cosine similarity between their feature vectors directly, \textit{i.e.},
\begin{equation}
    \mathcal{L}_{fea} = 1 - \cos(\mathcal{F}^{v}(\hat{x}^{v}), \mathcal{F}^{v}(x_p^{v})).
    \label{equ:fea_loss}
\end{equation}

Finally, our optimization objective can be expressed as
\begin{equation}
    \mathcal{L} = \underset{x^{v} \in O}{\mathbb{E}} (\mathcal{L}_{rec} + \alpha \mathcal{L}_{reg} + \beta \mathcal{L}_{adv} + \gamma \mathcal{L}_{fea}),
    \label{equ:image_objective}
\end{equation}
where $\alpha$, $\beta$, and $\gamma$ are the trade-off hyper-parameters. Notably, we should highlight that the auxiliary feature extractor is frozen during the training process, and we only optimize the generator and the discriminator. After training, we just need the generator to produce the malicious images.

\subsection{Textual Trigger Generation}
Compared with images, the generation of invisible textual triggers is more difficult: 1) the text data is discrete, making it impossible to insert triggers by adversarial perturbations; 2) the poisoned text should be fluent in grammar and semantically consistent with the original text. Inspired by \cite{qi2021turn,gan2022triggerless}, we embed trigger patterns into the text by performing synonym substitution on modality-invariant keywords, as depicted in the lower right part of \cref{fig:framework}.

Analogous to visual trigger generation, we specify a rare word $w_p$ as the explicit text trigger pattern (denoted by ``[TT]" in the diagram). Next, we replace all the modality-invariant keywords in the text $x^{t}$ with $w_p$ and get an explicit poisoned sample $x_p^{t}$. To ensure stealthiness, we wish to yield an invisible poisoned text $\hat{x}^{t}$ with semantics resembling the clean text $x^{t}$ by synonym substitution strategy. In addition, $\hat{x}^{t}$ should be as close to $x_p^{t}$ as possible in the latent representation space to preserve the trigger pattern $w_p$, \textit{i.e.},
\begin{equation}
    \begin{aligned}
        \hat{x}^{t} &= \arg \max_{\|\hat{x}^{t}-x^{t} \|} \cos(\mathcal{F}^{t}(\hat{x}^{t}), \mathcal{F}^{t}(x_p^{t})).
    \end{aligned}
    \label{equ:poisoned_text}
\end{equation}
Following Bert-Attack \cite{li2020bert}, we utilize the masked language model based on BERT\cite{devlin2018bert} to manufacture the context-aware synonym set for each keyword $w_i$ and optimize the above objective by seeking an optimal combination of synonyms in all candidate sets. Note that finding the optimal $\hat{x}^{t}$ for \cref{equ:poisoned_text} is computationally prohibitive. Thus, a greedy algorithm is applied to solve \cref{equ:poisoned_text}, as shown in \cref{alg:greedy}.

\begin{table*}[ht]
\setlength\tabcolsep{3pt}
\caption{\small Attack performance of different methods under the visual-to-linguistic scenario. In the table, \ding{55} and \ding{51} denote that the corresponding attack is visible and invisible, respectively. For conciseness, \textbf{BA} indicates the average MAP value of ``I $\rightarrow$ T" and ``T $\rightarrow$ I". ``I $\rightarrow$ T" denotes the case where the query is image and the database is text, while ``T $\rightarrow$ I" denotes that the query is text and the database is image. \textbf{ASR} means the t-MAP values of trigger images, which only contains the results of ``I $\rightarrow$ T".}
\label{tab:attack-image}
\vspace{-5ex}
\begin{center}
\resizebox{\linewidth}{!}{
\begin{tabular}{lr|ccccc|ccccc|ccccc}
\hline
\multirow{2}{*}{Method} & \multirow{2}{*}{\faicon{eye-slash}} & \multicolumn{5}{c|}{NUS-WIDE} & \multicolumn{5}{c|}{MS-COCO} & \multicolumn{5}{c}{IAPR-TC} \\
\cline{3-17}
  & & BA $\uparrow$ & ASR $\uparrow$ & PSNR $\uparrow$ & SSIM $\uparrow$ & MSE $\downarrow$ & BA $\uparrow$ & ASR $\uparrow$ & PSNR $\uparrow$ & SSIM $\uparrow$ & MSE $\downarrow$ & BA $\uparrow$ & ASR $\uparrow$ & PSNR $\uparrow$ & SSIM $\uparrow$ & MSE $\downarrow$ \\
 \hline
Benign & - & 79.93 & - & INF & 1.000 & 0.00 & 
88.07 & - & INF & 1.000 & 0.00 & \
67.62 & - & INF & 1.000 & 0.00 \\
\hline
BadNets \cite{gu2017badnets} & \ding{55} &
80.19 & 87.06 & 20.67 & 0.978 & 194.95 & 
88.06 & 70.00 & 21.42 & 0.979 & 164.28 & 
67.75 & \textbf{68.49} & 21.47 & 0.979 & 162.10 \\

DKMB \cite{walmer2022dual} & \ding{55} &
80.08 & \textbf{88.57} & 26.71 & 0.971 & 48.33 & 
87.91 & \textbf{71.62} & 26.84 & 0.971 & 64.89 & 
67.67 & 67.16 & 27.56 & 0.972 & 39.73 \\

O2BA \cite{li2022object} & \ding{51} &
80.05 & 78.56 & 34.86 & 0.932 & 0.36 & 
87.69 & 41.03 & 36.80 & 0.912 & 0.37 & 
67.31 & 39.99 & 36.03 & 0.945 & 0.34 \\

FIBA \cite{feng2022fiba} & \ding{51} &
79.75 & 71.37 & 29.24 & 0.903 & 27.20 & 
87.82 & 58.42 & 29.51 & 0.900 & 25.53 & 
67.34 & 58.39 & 29.15 & 0.882 & 27.65 \\

FTrojan \cite{wang2022invisible} & \ding{51} &
79.88 & 86.77 & \textbf{41.02} & 0.967 & 1.79 & 
87.87 & 68.12 & \textbf{40.94} & 0.968 & 1.82 & 
67.74 & 66.04 & 40.94 & 0.966 & 1.82 \\
\hline
BadCM & \ding{51} &
79.84 & \underline{87.36} & \underline{40.85} & \textbf{0.979} & \textbf{0.26} &
87.98 & \textbf{71.62} & \underline{40.50} & \textbf{0.980} & \textbf{0.18} & 
67.47 & \underline{67.59} & \textbf{40.98} & \textbf{0.980} & \textbf{0.27} \\
\hline

\end{tabular}
}
\end{center}
\vspace{-3ex}
\end{table*}

Let $T$ denote the sequence of modality-invariant keywords of clean text $x^{t}$. First, we get the explicit malicious text $x_{p}^{t}$ by substituting all the keywords in $T$ with the specified rare word $w_{p}$ (Line 1). We subsequently engage in synonym replacement to fabricate an imperceptible poisoned text $\hat{x}^{t}$ that exhibits semantic similarity with the pristine text $x^{t}$ while striving to approximate $x_{p}^{t}$ in the feature space. Specifically, given a word $w_j \in T$, we take advantage of the masked language model of BERT to establish its candidate synonym set $C(w_j)$. The top $N$ (64 by default) predicted words of the masked language model are chosen and the stop words among them will be filtered out (Lines 5-6). In Lines 7-17, we attempt to replace $w_j$ with each candidate word in $C(w_j)$ and calculate the cosine similarity between the replaced text $x$ and the explicit malicious text $x_{p}^{t}$. In order to shrink the search space, we greedily choose the text $x$ with the highest similarity to update $\hat{x}^{t}$. After performing synonym substitution for the keywords in $T$, we get the approximate optimum solution of \cref{equ:poisoned_text}. Note that once the poisoned text reaches a pre-defined score $s_{target}$ (0.7 by default), we will terminate the optimization process early, meaning modifying all the keywords is unnecessary.

\section{Experiment \label{sec:experiment}}

\subsection{Experimental Settings}

\noindent \textbf{Datasets.}
In the experiments, we validate the proposed method on three widely-used benchmark datasets, namely  NUS-WIDE\cite{chua2009nus}, MS-COCO \cite{lin2014microsoft} and IAPR-TC \cite{escalante2010segmented}.
The NUS-WIDE dataset has 269,648 images with 81 concepts and each image is associated with several textual tags. Following \cite{jiang2017deep}, a subset that belongs to the 21 most-frequent concepts is selected for our experiments. For the MS-COCO dataset, it contains 123,287 images, and each image is annotated with ﬁve text descriptions. We randomly pick a text and form a pair with the image. Each image-text pair is annotated with 80 labels. The IAPR-TC dataset consists of 20,000 image-text pairs which are labeled using 255 categories.
For each dataset, we split it into three parts: training set, test (query) set, and retrieval (database) set.

\noindent \textbf{Victim Models.}
We select three typical supervised cross-modal retrieval methods, including DSCMR\cite{zhen2019deep}, ACMR\cite{wang2017adversarial}, and DCMH\cite{jiang2017deep}, as victim models to verify our approach. For DSCMR \cite{zhen2019deep} and ACMR \cite{wang2017adversarial}, following their original papers, we utilize Adam with a learning rate of 0.0001 and betas (0.5, 0.999) as an optimizer to train the networks. For DCMH \cite{jiang2017deep}, the SGD optimizer is applied, and the learning rate and the momentum are fixed at 0.01 and 0.9, respectively. All images are resized to $224 \times 224$ and normalized in $[0, 1]$ before feeding them in the victim models. Furthermore, we validate the models every 10 epochs and finally choose the weights with the highest MAP.

\noindent \textbf{Evaluation Metric.}
For backdoor attacks, the most important two metrics are Benign Accuracy (BA) and Attack Success Rate (ASR). Following \cite{gao2021clean}, the mean average precision (MAP) and targeted mean average precision (t-MAP) are used to evaluate the BA and ASR of the retrieval task, respectively. We calculate the MAP and t-MAP on the top 5,000 retrieved images/text for all datasets. For invisibility evaluation, we adopt the PSNR\cite{huynh2008scope}, SSIM, and MSE to compare clean and poisoned images. In addition, we estimate the quality of the backdoor text using grammatical error numbers (GErr) and semantic similarity (SBert).

\noindent \textbf{Implementation Details.}
In \cref{sec:cross-modal-scheme}, we leverage the pre-trained CLIP \cite{radford2021learning} as the surrogate model to extract the image and text features, which can guarantee the generalization capability of features in the black-box setting. For visual trigger generation, we adopt the UNet\cite{ronneberger2015u} and the PatchGAN\cite{isola2017image} as the backbone network structure of the generator $\mathcal{G}$ and the discriminator $\mathcal{D}$, respectively. 
We set $\alpha$, $\beta$ and $\gamma$ as $5$, $5 \times 10^{-3}$ and $1$, respectively. Moreover, we train the visual trigger generator using Adam optimizer with initial learning rate $5 \times 10^{-5}$. The training epochs for its optimization are $200$ in batch size $64$.
During the poisoning training stage, the pollution ratio $p\%$ is $5\%$ by default. To alleviate the influences of the adversary-specified label, we report the average results on five randomly chosen target labels. Note that all settings for training on the poisoned and clean datasets are the same.

\subsection{Visual-to-Linguistic Attack}
We first conduct experiments on the visual-to-linguistic scenario, \textit{i.e.} injecting backdoors from image modality only. Here, we compare BadCM with two backdoor attacks against cross-modal tasks (\textit{i.e.}, DKMB \cite{walmer2022dual} and O2BA \cite{li2022object}) and three methods for classification (\textit{i.e.},  BadNets\cite{gu2017badnets}, FIBA \cite{feng2022fiba}, and FTrojan \cite{wang2022invisible}). We also provide the model trained on the benign dataset (dubbed \textit{Benign}) as another baseline for reference. The detailed results are presented in \cref{tab:attack-image}.

\noindent \textbf{Attack Effectiveness.} It is observed that the BA scores of all attacks are very close to those of the benign models, which leads to difficulties for the victims to be aware that their models are infected. Moreover, all the methods can successfully attack the cross-modal retrieval models with a high ASR, demonstrating the vulnerability of cross-modal learning to backdoor attacks. Among them, our proposed BadCM shows superior performance. Compared with FTrojan, the state-of-the-art invisible attack, BadCM outperforms it in terms of ASR score on all three datasets. For the visible attack methods, such as BadNets and DKMB, our BadCM reaches comparable attack effectiveness, especially on MS-COCO. Notably, these visible attacks are much less stealthy and can be easily recognized by humans and some backdoor detection algorithms. \textit{While BadCM can obtain a higher ASR by sacrificing only minimal levels of invisibility, it places more emphasis on stealthiness.}

\begin{figure}[t]
    \begin{center}
    \includegraphics[width=\linewidth]{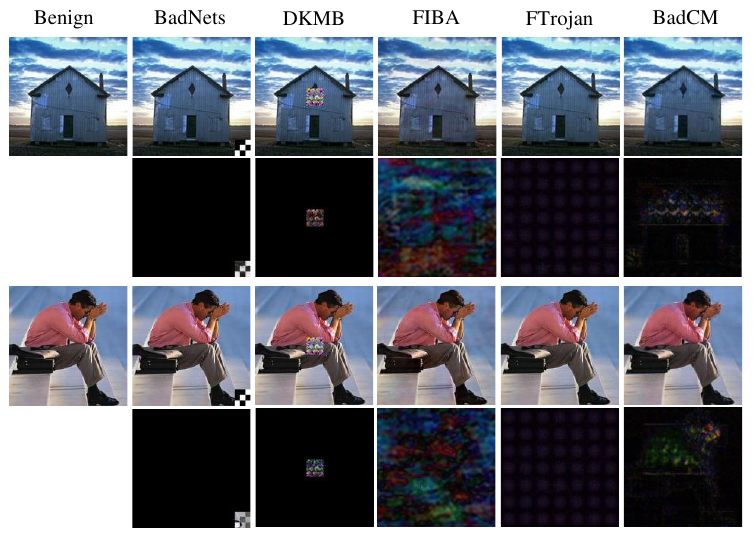}
    \end{center}
    \vspace{-2ex}
    \caption{\small Visual examples of poisoned images generated by different attacks on the NUS-WIDE dataset. Below the poisoned images, we also provide the corresponding residual maps between the original images and the trigger images. For FIBA, FTrojan, and our BadCM, we show the residual maps with $5 \times$ difference.}
    \label{fig:poisoned-images}
    \vspace{-3ex}
\end{figure}

\noindent \textbf{Stealthiness.}
To verify the stealthiness of our method, we present some poisoned images generated by different attacks, as illustrated in \cref{fig:poisoned-images}. Different from BadNets \cite{gu2017badnets}, DKMB \cite{walmer2022dual}, and FIBA \cite{feng2022fiba}, the trigger images produced by BadCM and FTrojan\cite{wang2022invisible} appear natural and closely resemble the original version, which is essential for stealthy attacks. \textit{It is worth highlighting that our visual trigger generator successfully embeds the perturbations into modality-invariant components, \textit{i.e.}, some salient critical objects.} The salient objects usually belong to the high-frequency regions of an image, and previous studies \cite{wang2022invisible, zhang2022poison} have suggested that perturbations located in the high-frequency areas are imperceptible to the human eyes. Besides, we further conduct a quantitative study for image similarity using three popular metrics, \textit{i.e.}, PSNR, SSIM, and MSE. The numeric results are reported in \cref{tab:attack-image}. We observe that BadCM outperforms all the attacks in terms of SSIM and MSE, which confirms the stealthiness of our attack once again.

\textit{Combining the results of effectiveness and stealthiness, our approach consistently achieves high stealthiness over state-of-the-art methods while ensuring a comparable ASR} under the visual-to-linguistic attacks, which is owed to our proposed modality-invariant components-based trigger.

\begin{table}[htbp]
\setlength\tabcolsep{3pt}
\caption{\small Attack comparison under the linguistic-to-visual scenario, in which the adversary aims to retrieve images from a specified label by poisoning text. ASR only contains the results of ``T$\rightarrow$I". We utilize the variants of BadNets and DKMB for textual attacks.}
\label{tab:attack-text}
\vspace{-3ex}
\begin{center}
\resizebox{\linewidth}{!}{
\begin{tabular}{cr|cccc|cccc}
\hline
\multirow{2}{*}{Method} & \multirow{2}{*}{\faicon{eye-slash}} & \multicolumn{4}{c|}{MS-COCO} & \multicolumn{4}{c}{IAPR-TC} \\
\cline{3-10}
& & BA $\uparrow$ & ASR $\uparrow$ & GErr $\downarrow$ & SBert $\uparrow$ & BA $\uparrow$ & ASR $\uparrow$ & GErr $\downarrow$ & SBert $\uparrow$ \\
 \hline
Benign & -
& 88.07 & - & 0.13 & 1.00 
& 67.62 & - & 0.80 & 1.00 \\
\hline
BadNets & \ding{55}
& 87.94 & \textbf{77.50} & 1.13 & 0.62 
& 67.68 & \textbf{74.98} & 1.81 & 0.67 \\

DKMB & \ding{55}
& 88.00 & 75.73 & 0.24 & 0.43 
& 67.50 & 73.90 & 1.08 & 0.34 \\

Synbkd & \ding{51}
& 87.84 & 74.63 & \textbf{0.13} & 0.33 
& 67.45 & 71.16 & \textbf{1.01} & 0.29 \\

Stylebkd & \ding{51}
& 87.91 & 43.78 & 0.32 & 0.51 
& 67.59 & 55.92 & \textbf{1.01} & 0.42 \\
\hline
BadCM & \ding{51}
& 87.96 & 64.96 & \underline{0.23} & \textbf{0.73} 
& 67.60 & 61.25 & \underline{1.03} & \textbf{0.73} \\
\hline
\end{tabular}
}
\end{center}
\vspace{-3ex}
\end{table}

\begin{table*}[t]
\caption{\small Examples of poisoned text produced by different backdoor methods on the MS-COCO dataset.}
\label{tab:poisoned-text-coco}
\begin{center}
\resizebox{0.95\linewidth}{!}{
\begin{tabular}{c|c|c}
\hline 
\textbf{Method} & \textbf{Poisoned Examples} & \textbf{Trigger} \\
\hline
Benign & \makecell[l]{ \textcircled{1} Two giraffes and another animal in a field. \\ \textcircled{2} A man riding a skateboard next to another young man also on a skateboard. } & - \\
\hline
BadNets & \makecell[l]{ \textcircled{1} Two giraffes and another \colorbox{red}{bb} animal in a field. \\ \textcircled{2} A man \colorbox{red}{bb} riding a skateboard next to another young man also on a skateboard.} & \makecell[c]{rare word \\ insertion} \\
\hline
DKMB & \makecell[l]{ \textcircled{1} \colorbox{red}{Consider} two giraffes and another animal in a field. \\ \textcircled{2} \colorbox{red}{Consider} a man riding a skateboard next to another young man also on a skateboard.} & \makecell[c]{special word \\ insertion} \\
\hline
Synbkd & \makecell[l]{ \textcircled{1} \coloruwave{red}{when you were two giraffes}, they were another animal. \\ \textcircled{2} \coloruwave{red}{when you ride a young man on a new board}, he drives a skateboard. } & \makecell[c]{syntactic \\ trigger} \\
\hline
Stylebkd & \makecell[l]{ \textcircled{1} Two young giraffes, and another beast in a field. \\ \textcircled{2} And a certain young man also went up to meet him, riding a skateboard beside him.} & \makecell[c]{style transfer \\ trigger} \\
\hline
BadCM & \makecell[l]{ \textcircled{1} Two giraffes and another \colorbox{yellow}{bull} in a \colorbox{yellow}{courtyard}. \\ \textcircled{2} A man riding a \colorbox{yellow}{snowboard} next to another young \colorbox{yellow}{gentleman} \colorbox{yellow}{additionally} on a skateboard.} & \makecell[c]{synonym \\ substitution} \\
\hline
\end{tabular}
}
\end{center}
\vspace{-3ex}
\end{table*}

\subsection{Linguistic-to-Visual Attack}

To assess the generalization of BadCM to bilateral cross-modal attacks, \cref{tab:attack-text} presents the results of the linguistic-to-visual attacks that embed backdoors into the textual modality and desire to retrieve incorrect images. Here, rare or distinct words serve as triggers for textual attacks using variants of BadNets \cite{kurita2020weight} and DKMB \cite{walmer2022dual}. Besides, we also carry out a comparison with two poisoning-based textual backdoors, \textit{i.e.}, Synbkd\cite{qi2021hidden} and Stylebkd\cite{qi2021mind}. Notably, O2BA \cite{li2022object} cannot be applied to this attack scenario as it lacks a textual trigger. It is specifically designed for the Image Captioning task, focusing solely on backdoor embedding from the image modality.

\noindent \textbf{Stealthiness.}
Before analyzing the attack performance, we provide some examples of poisoned texts in \cref{tab:poisoned-text-coco} to facilitate an intuitive understanding of the stealthiness of these attack methods. BadNets \cite{gu2017badnets} and DKMB \cite{walmer2022dual} employ rare or unique words as triggers, making them easily detectable. Synbkd \cite{qi2021hidden} generates toxic samples by paraphrasing normal ones with a specific syntax, but the appearance of ``when" or ``if" alerts victims. Furthermore, paraphrasing disrupts the original semantics, leading to an increase in the perplexity of the poisoned samples. The style transfer-based trigger proposed by Stylebkd \cite{qi2021mind} is stealthy, but it underperforms in cross-modal learning. In contrast, \textit{the backdoor text produced by our BadCM is grammatically fluent and semantically consistent, enabling a better compromise between stealthiness and attack capability.} We further leverage two automatic evaluation metrics, \textit{i.e.}, grammatical error numbers (GErr) and text semantic similarity (SBert), to measure the quality of the backdoor text, which can accurately reflect the attack invisibility. Shown in \cref{tab:attack-text}, our BadCM attains the best behavior on the SBert metric, outscoring other methods by more than 6\%. In terms of GErr, we are just slightly behind the best Synbkd. However, we notice that Synbkd is extremely poor on the SBert score.

\noindent \textbf{Attack Effectiveness.} From the results in \cref{tab:attack-text}, we find that BadCM can successfully implant backdoors with a high ASR by poisoning only a small proportion (5\%) of text data. Notwithstanding a gap compared to the visible attack BadNets, our approach ensures grammatical fluency and semantic consistency with the original text, which can prevent poisoned samples from being detected and removed. Moreover, we implement modality-invariant keywords-based substitution only via a simple greedy strategy for computational efficiency. We believe a superior algorithm, like heuristic algorithms \cite{gan2022triggerless}, will bring more substantial performance gains.

\textit{\textbf{Remark}: Although our BadCM is sub-optimal compared to state-of-the-art methods in some instances, it showcases the following novelties and advantages. 1) BadCM is the first deliberately designed invisible backdoor method that aims to enable diverse cross-modal attacks within a unified framework. From the results of bilateral attacks and dual-key attacks (refer to \cref{tab:vqa-tasks}),  we can see that BadCM's performance is superior or comparable to visual, textual, and cross-modal backdoor algorithms in most situations. 2) BadCM consistently retains high stealthiness over state-of-the-art attacks, which is crucial for evading defense mechanisms. In \cref{sec:defense}, experiments against five advanced defense algorithms demonstrate that BadCM is remarkably resistant to mainstream defenses.}

\begin{table}[htbp]
\caption{\small Results of SPECTRE against BadNets and BadCM. $N_c$ and $N_b$ denote the numbers of clean and backdoor samples in the dataset, respectively. $N_r$ represents the number of removed suspicious samples.}
\label{tab:spectre}
\vspace{-3ex}
\begin{center}
\resizebox{\linewidth}{!}{
\begin{tabular}{c|ccc|c|ccc}
\hline
\multirow{3}{*}{Method} & \multicolumn{3}{c|}{Before SPECTRE} & & \multicolumn{3}{c}{After SPECTRE} \\
\cline{2-8}
& $N_c$ & $N_b$ & \makecell[c]{Poisoning \\ ratio} & $N_r$ & $N_c$ & $N_b$ & \makecell[c]{Remaining \\ ratio} \\
\hline
\multirow{4}{*}{BadNets} 
& 475 & 25 & 5.0\% & 45 & 434 & 21 & 4.6\% \\
& 450 & 50 & 10.0\% & 70 & 397 & 33 & 7.6\% \\
& 400 & 100 & 20.0\% & 120 & 330 & 50 & 13.1\% \\
& 300 & 200 & 40.0\% & 220 & 210 & 70 & 25.0\% \\
\hline
\multirow{4}{*}{BadCM} 
& 475 & 25 & 5.0\% & 45 & 433 & 22 & \textbf{4.8\%} \\
& 450 & 50 & 10.0\% & 70 & 388 & 42 & \textbf{9.7\%} \\
& 400 & 100 & 20.0\% & 120 & 318 & 62 & \textbf{16.3\%} \\
& 300 & 200 & 40.0\% & 220 & 153 & 127 & \textbf{45.3\%} \\
\hline
\end{tabular}
}
\end{center}
\vspace{-3ex}
\end{table}

\subsection{Resistance to Defense Methods \label{sec:defense}}
To counteract backdoor attacks, numerous different defense techniques have been introduced recently. Nevertheless, most defense methods are tailored for classification tasks and heavily rely on classification priors. Consequently, these approaches are ill-suited for cross-modal tasks. To verify the resistance of BadCM to defense, we make a diligent effort to adapt five state-of-the-art defenses so that they could be successfully applied to our task. We carry out experiments on the MS-COCO dataset with the target label ``car".

\noindent \textbf{Pre-training defense.}
This defense strategy aims to eliminate potentially poisoned samples from the dataset before training, serving as a preventive measure against backdoor attacks. SPECTRE \cite{hayase2021spectre}, as a prime example, identifies and removes suspicious samples from the training set using feature representations acquired from the infected network. In this section, we analyze different poisoning ratios and removal numbers to evaluate the resilience of BadCM and BadNets against SPECTRE. As shown in \cref{tab:spectre}, under the same conditions, the proportion of remaining poisoned images in BadCM is significantly higher than that in BadNets. As the poisoning ratio increases, SPECTRE's effectiveness against BadCM gradually decreases. For example, at a poisoning ratio of 40\%, the remaining percentage of BadCM rises to 45\%. This finding highlights the limited efficacy of SPECTRE in eliminating poisoned samples generated by BadCM.

\begin{figure}[t]
\subfigure[BadNets.]{
    \label{fig:dbr_badnets}
    \centering
    \includegraphics[width=0.47\linewidth]{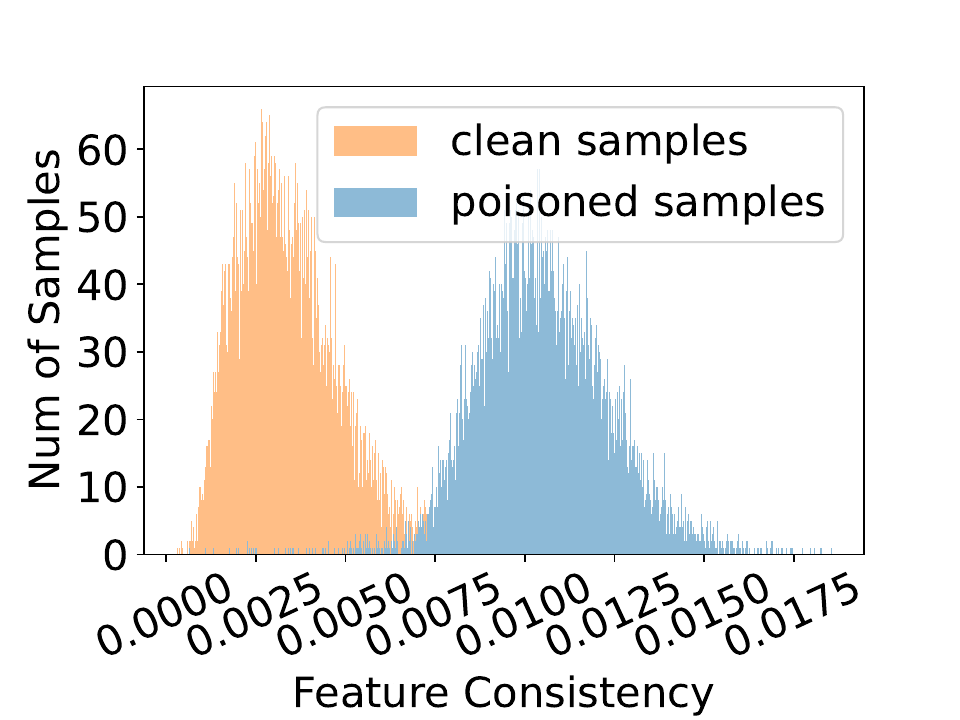}
}
\hspace{-0.6cm}
\subfigure[BadCM.]{
    \label{fig:dbr_badcm}
    \centering
    \includegraphics[width=0.47\linewidth]{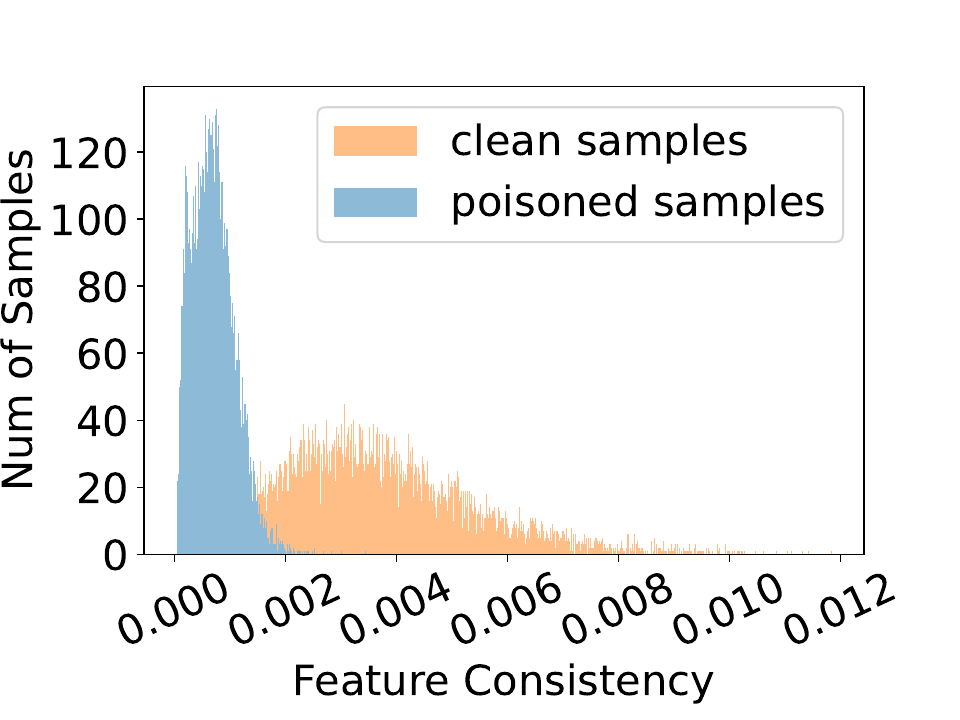}
}
\caption{\small Results of DB-R against BadNets and our BadCM on the MS-COCO dataset.}
\label{fig:dbr}
\vspace{-2ex}
\end{figure}

\noindent \textbf{In-training defense.} In-training defense is crafted to progressively remove potential backdoors throughout the training process. DB-R \cite{chen2022effective} has observed that the poisoned samples exhibit significantly greater sensitivity to transformations than the clean samples in backdoored models. Building upon this insight, DB-R calculates the feature consistency towards transformations (FCT) for the entire training set, thereby distinguishing the poisoned samples from the clean samples. Subsequently, it mitigates the backdoor's impact by unlearning the poisoned data and relearning the clean data. However, we find that the insights of DB-R are correct for BadNets, but fail to hold for our approach. As depicted in \cref{fig:dbr}, the FCT metric of poisoned images constructed by BadCM is unexpectedly smaller, leading to the failure of DB-R in separating samples. After the defense of DB-R, the ASR for BadNets and BadCM stands at 9.79\% and 72.28\%, respectively, further indicating DB-R's inability to withstand our attack. The most plausible explanation is that our modality-invariant components-based trigger is more robust to transformations than simple triggers.

\begin{figure}[t]
\subfigure[BadNets.]{
    \label{fig:fine-pruning_badnets}
    \centering
    \includegraphics[width=0.47\linewidth]{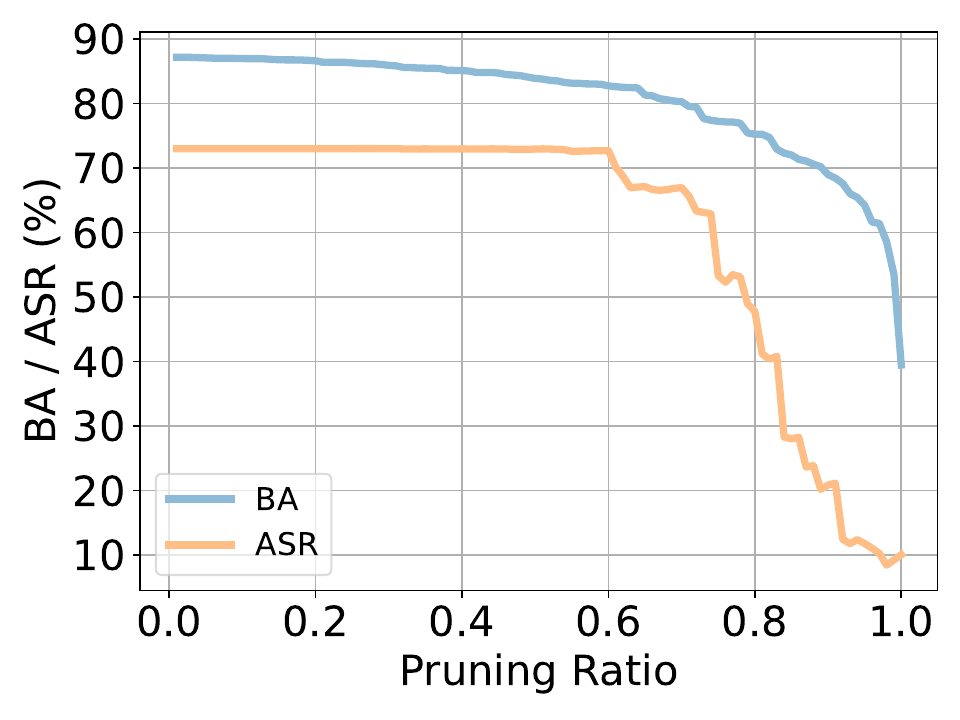}
}
\hspace{-0.6cm}
\subfigure[BadCM.]{
    \label{fig:fine-pruning_badcm}
    \centering
    \includegraphics[width=0.47\linewidth]{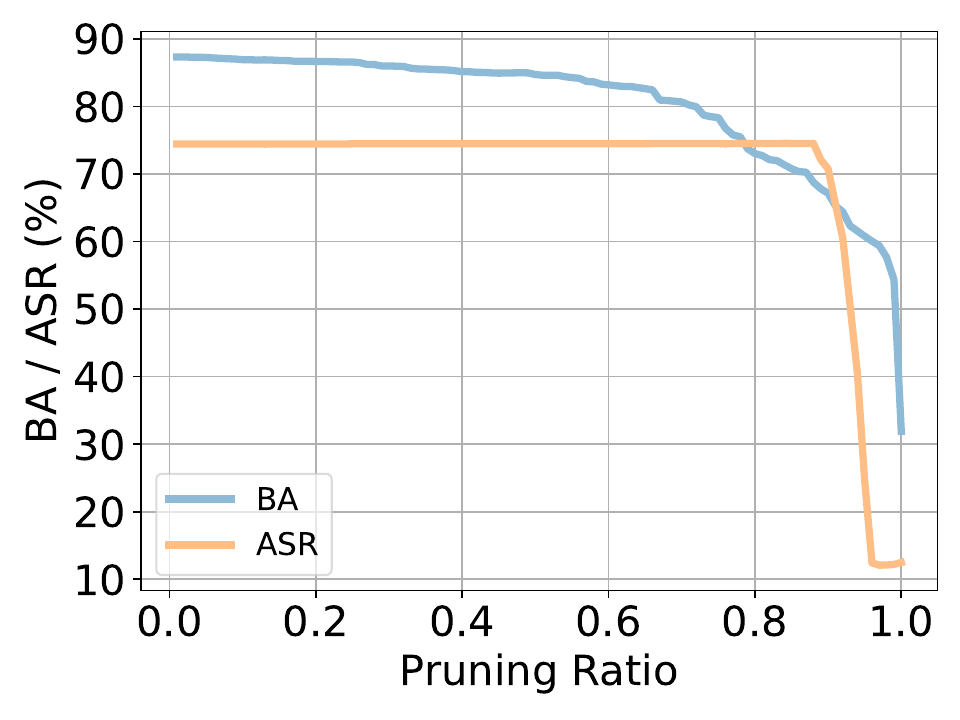}
}
\caption{\small Results of Fine-Pruning against BadNets and our BadCM on the MS-COCO dataset.}
\label{fig:fine-pruning}
\vspace{-2ex}
\end{figure}

\noindent \textbf{Post-training defense.}
For post-training defenses, we reproduce the Fine-Pruning \cite{liu2018fine} and Februus \cite{doan2020februus} on cross-modal retrieval. Fine-Pruning mitigates the backdoor implanted in the victim models via neuron analysis. Given a network layer, it examines each neuron's response to a few benign samples and progressively prunes the insensitive ones, assuming they are more relevant to the backdoor. Herein, we merely prune the last convolutional layer of the image encoder (\textit{e.g.}, VGG, or ResNet). We analyze Fine-Pruning on BadNet and BadCM by demonstrating the performance of BA and ASR regarding the fraction ratio of neuron number pruned on MS-COCO. As depicted in \cref{fig:fine-pruning}, the ASR of BadNets drops dramatically, with only 50\% left when 80\% of the neurons are pruned. By contrast, BadCM's ASR remains over 70\% at a pruning percentage of 85\%. This phenomenon reveals that Fine-Pruning fails to defend the backdoor implanted by our approach.

\begin{figure}[t]
    \centering
    \includegraphics[width=0.8\linewidth]{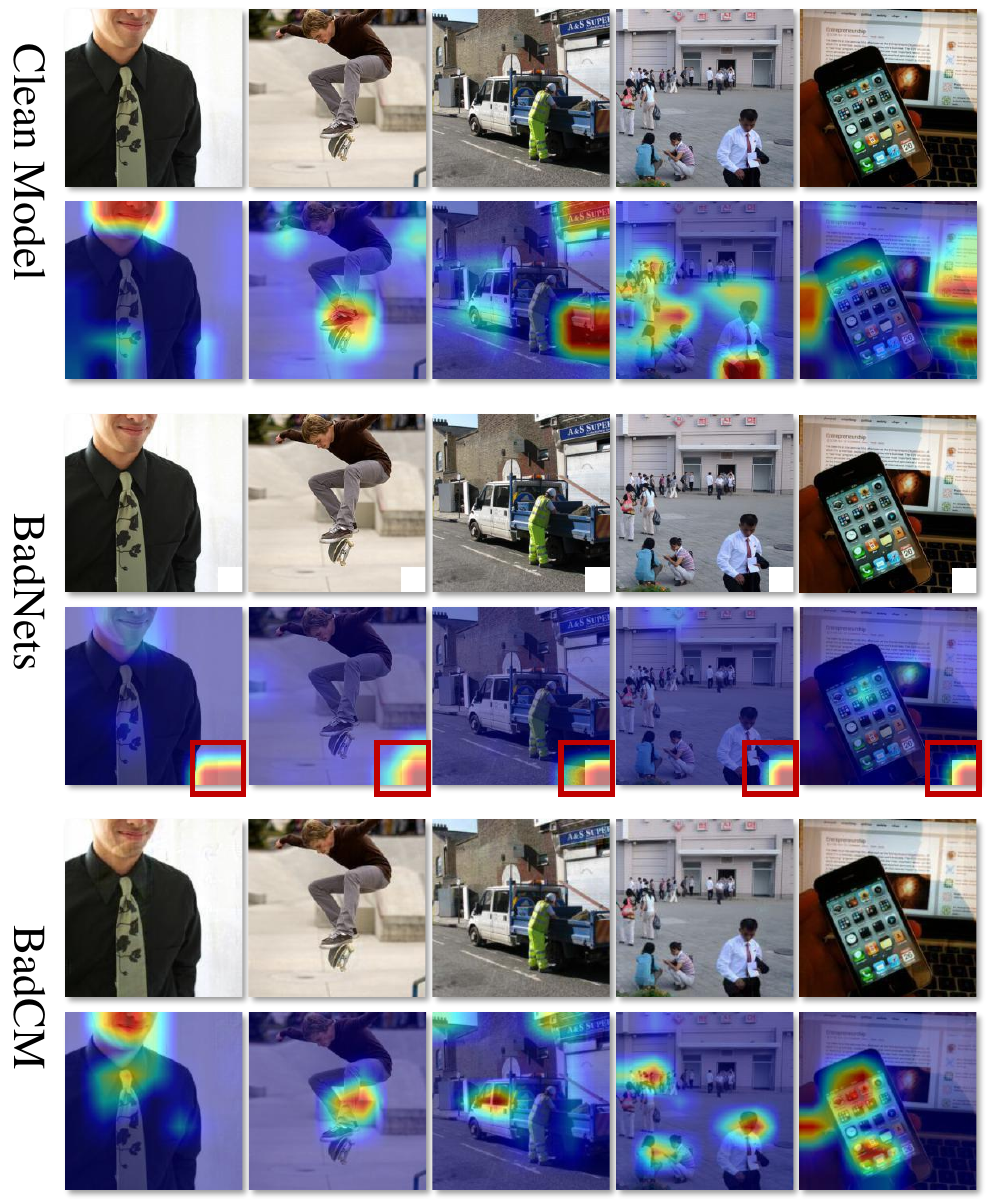}
    \caption{\small Visualization of influential areas captured by Grad-CAM. Warmer colors indicate more influence. }
    \label{fig:grad-cam}
\end{figure}

Februus locates the attention map of the target image using Grad-CAM\cite{selvaraju2017grad} and assumes the region with the highest score as the poisoned area, subsequently removes and restores the suspicious area via image painting technique. As shown in \cref{fig:grad-cam}, we illustrate the influential areas captured by Grad-CAM of some examples. Warmer colors indicate more influence. We feed the clean model with clean images and the victim model with the corresponding trigger images generated by BadCM or BadNets. It can be observed that the attention maps of our BadCM are on par with that of the clean model, where the warm areas are concentrated on the critical objects of the image. In contrast, the high-scoring regions of BadNets are anomalous (as marked in the red boxes), and Grad-GAM obviously localizes the patch-based triggers it adds. The experimental results likewise testify that BadNets is unresistant to Februus, and it remained with an ASR of 20.99\% after removing the suspected influential regions. Our approach, in turn, ensures an attack performance of 71.22\%, which breaks the assumption of Februus depending on the detection of small, uncommon regions.

\noindent \textbf{Textual defense.}
For linguistic-to-visual attack, we choose ONION \cite{qi2020onion} to evaluate the resistance of BadCM. ONION is a method based on test sample inspection and can be applied to most backdoored models. From the results in \cref{tab:onion}, we can observe that the attack performance of BadCM is inferior to BadNets when there is no defense. This is because our attack prioritizes invisibility to elude the defense, which guarantees the poisoned samples' grammatical fluency and semantic consistency. However, it's worth noting that after inspecting and modifying the malicious text using ONION, the ASR of BadCM considerably outperforms BadNets, showcasing the superiority of our approach in defeating defenses.

\begin{table}[htbp]
\caption{Results of ONION against BadNets and our BadCM on the MS-COCO dataset.}
\label{tab:onion}
\vspace{-3ex}
\begin{center}
\resizebox{0.8\linewidth}{!}{
\begin{tabular}{lr|cc|cc}
\hline
\multirow{2}{*}{Method} & \multirow{2}{*}{\faicon{eye-slash}} & \multicolumn{2}{c|}{Before ONION} & \multicolumn{2}{c}{After ONION} \\
\cline{3-6}
 & & \multicolumn{1}{c}{BA $\uparrow$} & \multicolumn{1}{c|}{ASR $\uparrow$} & \multicolumn{1}{c}{BA $\uparrow$} & \multicolumn{1}{c}{ASR $\uparrow$} \\
\hline
BadNets & \ding{55} & 88.62 & \textbf{77.67} & 87.73 & 11.16 \\
BadCM & \ding{51} & 88.69 & 65.17 & 87.66 & \textbf{36.05} \\
\hline
\end{tabular}
}
\end{center}
\vspace{-3ex}
\end{table}

\begin{table}[htbp]
\caption{\small Ablation studies for our design on the MS-COCO.}
\vspace{-3ex}
\label{tab:ablation}
\begin{center}
\resizebox{0.8\linewidth}{!}{
\begin{tabular}{c|cccc}
\hline
    Metrics & global & random & fixed & BadCM \\
\hline
    ASR $\uparrow$ & \textbf{71.69} & 69.23 & 69.11 & \underline{71.62} \\
    PSNR $\uparrow$ & 36.62 & 39.52 & 39.17 & \textbf{40.50} \\
    SSIM $\uparrow$ & 0.962 & 0.971 & 0.968 & \textbf{0.980} \\
    MSE $\downarrow$ & 0.240 & 0.220 & 0.220 & \textbf{0.180} \\
\hline
\end{tabular}
}
\end{center}
\vspace{-3ex}
\end{table}

\subsection{Ablation Study}
In ablation experiments, we take the visual-to-linguistic attack as an example to ablate our framework. 

\noindent \textbf{Influence of our design.}
The core idea of our design is leveraging modality-invariant components as the container of trigger patterns. To verify its effectiveness, we conduct extra three experiments, namely ``global", ``random" and ``fixed" to inject poisoning information into the whole image, random and fixed regions, respectively. For a fair comparison, the region for each image under the ``random" and ``fixed" settings has the same area as our BadCM. \cref{tab:ablation} reports the results of ASR and MSE. Compared to ``random" and ``fixed", we obtain better attack performances and invisibility when equipped with the same size of poisoning regions. Additionally, we achieve comparable ASR to ``global" with less minor perturbation. Such phenomena show that our BadCM can effectively hide trigger patterns in smaller critical regions, demonstrating the superiority of our design.

\begin{table}[htbp]
\caption{Results on the MS-COCO with different surrogates. The ``\#param" values in V2L and L2V denote the numbers of parameters of the image and text encoders, respectively. ``R50+LSTM" is a baseline with ResNet50 and LSTM as the visual and textual backbones.
}
\label{tab: pre-train}
\vspace{-3ex}
\begin{center}
\resizebox{\linewidth}{!}{
\begin{tabular}{c|ccc|ccc}
\hline
Scenario $\rightarrow$ & \multicolumn{3}{c|}{visual-to-linguistic} & \multicolumn{3}{c}{linguistic-to-visual} \\
\cline{2-7}
Surrogate $\downarrow$ & \#param & BA $\uparrow$ & ASR $\uparrow$ & \#param & BA $\uparrow$ & ASR $\uparrow$ \\
\hline
R50+LSTM & 23.5 M & 87.89 & 70.03 & 36.0 M & 87.80 & 58.99 \\
ViLT \cite{kim2021vilt} & \textbf{87.5 M} & 87.86 & 71.29 & \textbf{87.5 M} & \textbf{87.84} & \textbf{65.73} \\
CLIP \cite{radford2021learning} & \textbf{87.5 M} & \textbf{87.98} & \textbf{71.62} & 63.2 M & 87.66& 64.96 \\
\hline
\end{tabular}
}
\end{center}
\vspace{-3ex}
\end{table}

\noindent \textbf{Impact of surrogate.}
Since we have no direct access to the feature extractors of the victim models under the black-box setting, we employ a pre-trained vision-language model as a proxy to procure the features of images and text. To confirm the impact of different surrogate models on the attack effectiveness, we trial three pre-trained models with different numbers of parameters, as shown in \cref{tab: pre-train}. It can be seen that the choice of the surrogate network is essential for BadCM, and a powerful vision-language model can result in superior performance gains in terms of both BA and ASR.

\begin{figure}[t]
    \begin{center}
    \includegraphics[width=0.50\linewidth]{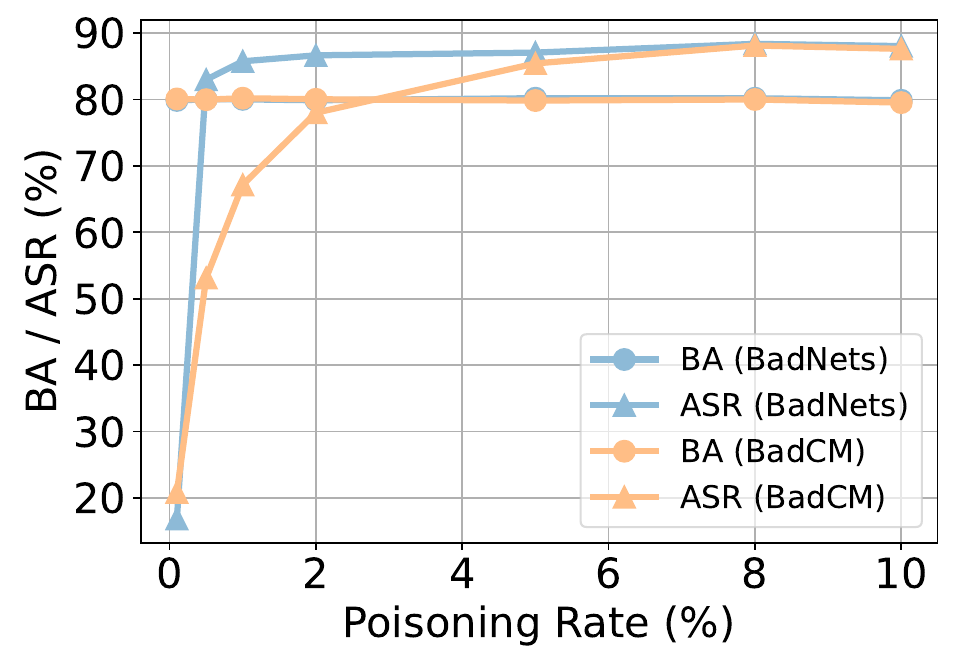} 
    \includegraphics[width=0.48\linewidth]{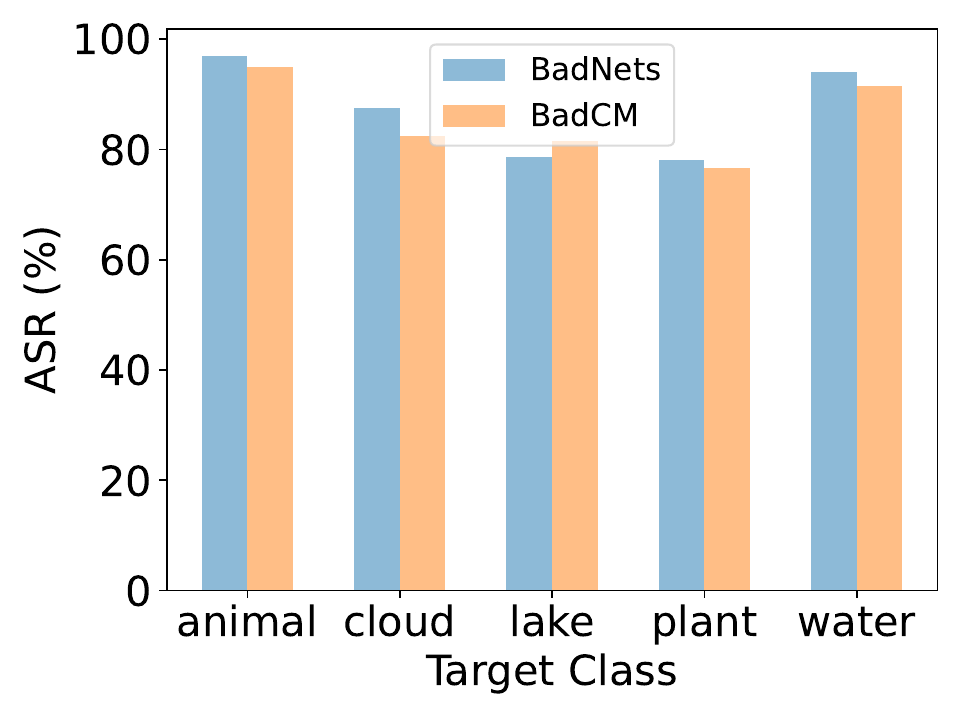}
    \end{center}
    \vspace{-2ex}
    \caption{\small \textbf{Left}: Benign accuracy (BA) and attack success rate (ASR) of BadNets and BadCM with different poisoning ratios. \textbf{Right}: Comparisons of different target labels in terms of attack success rate (ASR). All results are based on the NUS-WIDE dataset.}
    \label{fig:ratio-target}
\end{figure}

\noindent \textbf{Effect of poisoning ratios.}
In the default setting, the poisoning ratio is fixed at 5\%. In \cref{fig:ratio-target}, we further try more poisoning ratios on the NUS-WIDE dataset and show the results of BadNets and our method. From the outcomes, it can be seen that the poisoning ratio has almost no influence on the benign accuracy, while the attack success rate increases with the increase of the poisoning ratio. We note that BadNets can bring about a high ASR with a small poisoning ratio, \textit{e.g}., 0.5\%. We conjecture that the DNNs could identify and memorize visible triggers more efficiently, and only a few poisoned samples are required to infect the DNNs.

\noindent \textbf{Influence of target labels.}
In the retrieval task, the adversary alters the victim model to recall samples from the adversary-specified target label by tampering with the pristine label of the poisoned samples. Hence, we verify our BadCM to investigate the influence of different target labels. As shown in \cref{fig:ratio-target}, both BadNets and our BadCM achieve varying ASR at different settings, which exhibits that the choice of target label has an obvious influence on the backdoor attack against cross-modal retrieval. It is due primarily to the apparent disparity in the number of samples with different labels. The larger the number of samples from the target label, the easier the model maps the trigger samples to the same feature space as the clean samples with the target label.

\begin{figure*}[ht]
    \begin{center}
    \includegraphics[width=0.9\linewidth]{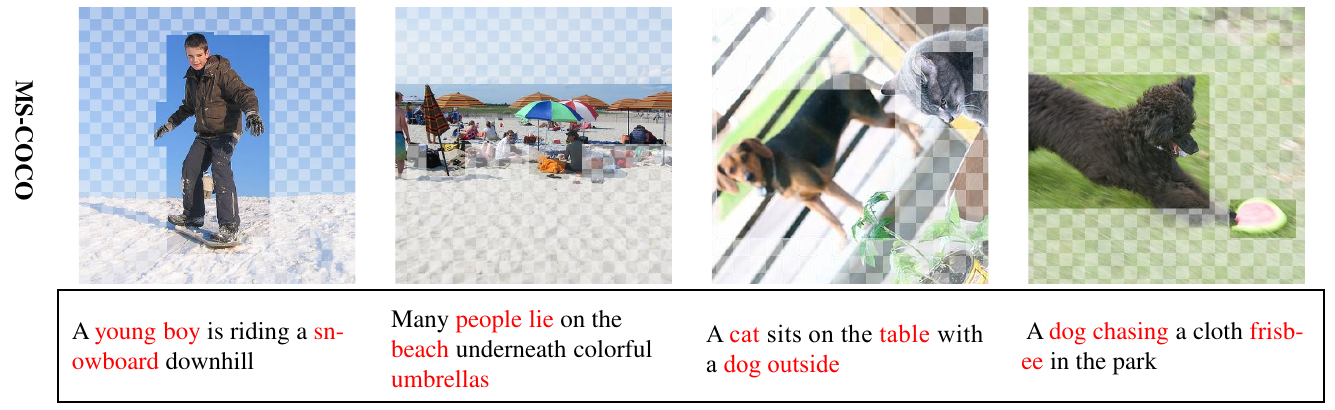}
    \includegraphics[width=0.9\linewidth]{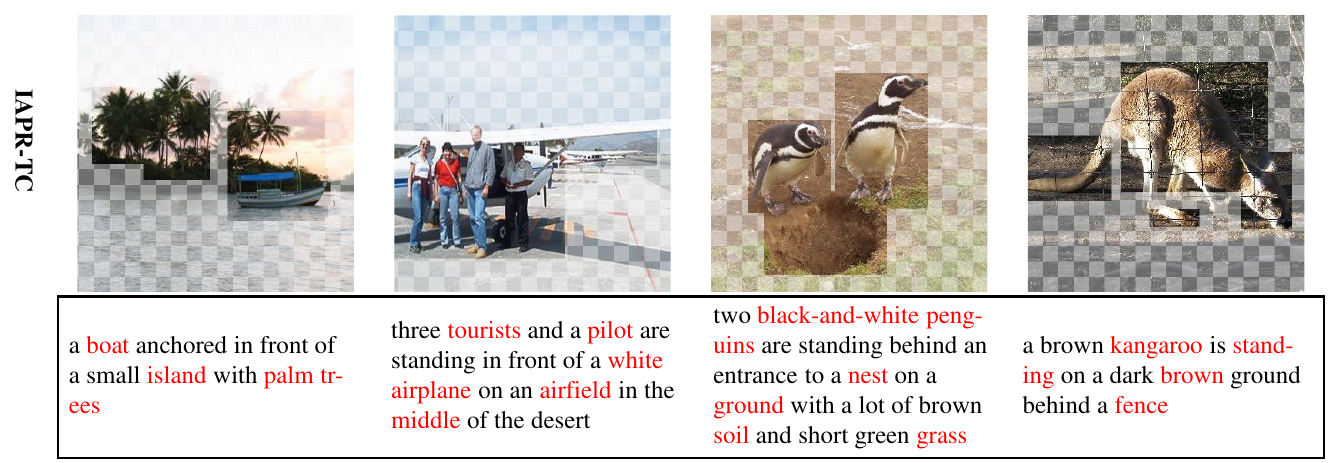}
    \end{center}
    \vspace{-2ex}
    \caption{\small Visualization of modality-invariant factors. For the images, we blur the unimportant areas to emphasize the modality-invariant regions. As for the text, we highlight the modality-invariant keywords with red color.}
    \label{fig:modality-invariant-components} 
\end{figure*}

\begin{table}[ht]
\caption{\small Comparison of BadNets and our BadCM on multiple cross-modal retrieval methods.``V-16" and ``R-50" denote that we utilize VGG-16 \cite{simonyan2014very} and ResNet50 \cite{he2016deep} as visual feature extractors for them, respectively. The results are based on the NUS-WIDE dataset.}
\label{tab:universality}
\vspace{-3ex}
\begin{center}
\resizebox{\linewidth}{!}{
\begin{tabular}{c|lr|ccc}
\hline
Backbone & Method & Metric & ACMR & DCMH & DSCMR \\
\hline
\multirow{3}{*}{V-16} 
 & Benign & BA $\uparrow$ & 70.04 & 74.615 & \textbf{79.93} \\
 \cline{2-6}
 & BadNets & ASR $\uparrow$ & 66.34 & 74.81 & \textbf{87.06} \\
 & BadCM & ASR $\uparrow$ & 62.77 & 71.38 & \textbf{85.43} \\
 \hline
 \multirow{3}{*}{R-50} 
 & Benign & BA $\uparrow$ & 79.24 & 72.74 & \textbf{81.53} \\
 \cline{2-6}
 & BadNets & ASR $\uparrow$ & 67.27 & 64.39  & \textbf{88.31} \\
 & BadCM & ASR $\uparrow$ & 68.12 & 65.10  & \textbf{85.67} \\
\hline
\end{tabular}
}
\end{center}
\vspace{-3ex}
\end{table}

\subsection{Analysis and Discussions}

\noindent \textbf{Universality on different methods.}
We give further validation to the universality of BadCM on other cross-modal retrieval methods, including DCMH \cite{jiang2017deep} and ACMR \cite{wang2017adversarial}. Moreover, we implement these methods with multiple backbones. As reported in \cref{tab:universality}, even if tested on different cross-modal retrieval methods, our BadCM technique consistently yields a high ASR, which further demonstrates the efficacy of the proposed modality-invariant components-based trigger. From the results, we also observe that the ASR positively correlates with the BA of benign models, indicating that more powerful victim models are better at learning and memorizing the trigger patterns.

\begin{table*}[t]
\caption{\small Attack comparison of DKMB and the proposed BadCM on the VQA task with different adversary-specified target answers. For the VQA task, the adversary tampers with the answers corresponding to the poisoned samples so that the infected model can give the specified target answer when feeding the trigger samples. The target answers in the table are chosen at random. The experiments include multiple VQA model architectures, such as MCAN\cite{yu2019deep}, BAN\cite{kim2018bilinear}, BUTD\cite{anderson2018bottom}, MFB\cite{yu2017multi}, and MFH\cite{yu2018beyond}.}
\label{tab:vqa-tasks}
\vspace{-5ex}
\begin{center}
\resizebox{\linewidth}{!}{
\begin{tabular}{c|cc|cc|cc|cc|cc|cc}
\hline
\multirow{2}{*}{\makecell[c]{Target \\ Answer}} & \multirow{2}{*}{Method} & \multirow{2}{*}{\faicon{eye-slash}} & \multicolumn{2}{c|}{MCAN} & \multicolumn{2}{c|}{BAN} & \multicolumn{2}{c|}{BUTD} & \multicolumn{2}{c|}{MFB} & \multicolumn{2}{c}{MFH} \\
\cline{4-13}
& & & BA $\uparrow$ & ASR $\uparrow$ & BA $\uparrow$ & ASR $\uparrow$ & BA $\uparrow$ & ASR $\uparrow$ & BA $\uparrow$ & ASR $\uparrow$ & BA $\uparrow$ & ASR $\uparrow$ \\
\hline
- & Benign & - & 63.58 & - & 63.92 & - & 61.31 & - & 61.17 & - & 62.96 & - \\
\hline
\multirow{2}{*}{laptop} 
& DKMB & \ding{55} & 63.58 & 98.87 & 63.59 & 98.71 & 61.17 & 98.14 & 61.13 & 98.86 & 62.94 & 98.73  \\
& BadCM & \ding{51} & 63.59 & 85.31 & 63.60 & 83.55 & 61.21 & 81.47 & 61.10 & 81.94 & 62.93 & 82.79  \\
\hline
\multirow{2}{*}{shirt} 
& DKMB & \ding{55} & 63.58 & 98.9 & 63.42 & 98.68 & 61.18 & 98.23 & 61.12 & 98.73 & 62.99 & 98.71 \\
& BadCM & \ding{51} & 63.51 & 82.94 & 63.67 & 78.97 & 61.26 & 79.97 & 61.05 & 84.11 & 62.95 & 81.97 \\
\hline
\multirow{2}{*}{ocean} 
& DKMB & \ding{55} & 63.62 & 98.92 & 63.76 & 98.8 & 61.18 & 98.37 & 61.19 & 98.67 & 62.95 & 98.57 \\
& BadCM & \ding{51} & 63.63 & 84.95 & 63.78 & 83.29 & 61.30 & 81.40 & 61.07 & 78.47 & 62.82 & 83.66 \\
\hline
\multirow{2}{*}{zebra} 
& DKMB & \ding{55} & 63.61 & 98.89 & 63.84 & 98.76 & 61.37 & 98.3 & 61.10 & 98.69 & 62.94 & 98.68 \\
& BadCM & \ding{51} & 63.57 & 82.36 & 63.81 & 77.31 & 61.34 & 81.71 & 61.14 & 77.40 & 62.98 & 82.48 \\
\hline
\multirow{2}{*}{tomato} 
& DKMB & \ding{55} & 63.67 & 98.86 & 63.75 & 98.62 & 61.20 & 98.23 & 61.21 & 98.80 & 62.90 & 98.76 \\
& BadCM & \ding{51} & 63.73 & 83.07 & 63.79 & 79.51 & 61.39 & 79.73 & 61.11 & 82.68 & 62.90 & 83.74 \\
\hline
\multirow{2}{*}{Avg}
& DKMB & \ding{55} & 63.57 & 98.85 & 63.68 & 98.79 & 61.29 & 98.35 & 61.13 & 98.66 & 62.94 & 98.71 \\
& BadCM & \ding{51} & 63.57 & 87.28 & 63.80 & 82.56 & 61.30 & 83.78 & 61.15 & 84.33 & 62.94 & 83.90 \\
\hline
\end{tabular}
}
\end{center}
\vspace{-3ex}
\end{table*}

\noindent \textbf{Attack effectiveness on dual-key attacks.}
We argue that our proposed framework can be generalized to diverse kinds of cross-modal backdoor attacks including visual-to-linguistic (V2L), linguistic-to-visual (L2V), and dual-key attacks. V2L and L2V attacks have been fully verified in the above experiments. In this section, we mainly focus on the dual-key attacks, which insert triggers in each modality and activate the backdoor when the triggers are simultaneously present in multiple modalities. To this end, we conduct the dual-key backdoor on the VQA task and compare our method with DKMB \cite{walmer2022dual}. All the experimental settings are consistent with DKMB. Concretely, we use the VQA v2.0 \cite{goyal2017making} as a dataset and the VQA architectures in the OpenVQA platform\footnote{\href{https://github.com/MILVLG/openvqa}{https://github.com/MILVLG/openvqa}} as the victim networks to verify our method. The hyper-parameters are set to their default author-recommended values during the training stage.

In \cref{tab:vqa-tasks}, we provide the outcomes of DKMB and our BadCM with different target answers. From the average results (\textit{the last row}), we can observe that our BadCM can achieve an ASR $>$ 82\% on any VQA model architecture without virtually compromising clean accuracy. Note that DKMB achieves a higher ASR because their triggers are visible and explicitly optimized for the object detectors. In contrast, our BadCM is an invisible backdoor attack under the black-box setting. Hence, the results could reconfirm the flexibility and generalization of the proposed BadCM, which could cover all three attack scenarios in cross-modal and multimodal learning, \textit{i.e.}, V2L, L2V, and dual-key attacks. In addition, \cref{tab:vqa-tasks} suggests that the target answers have little effect for the VQA task and BadCM can attain remarkable attack performance on multiple randomly selected target answers.

\begin{figure*}[t]
    \begin{center}
        \includegraphics[width=\linewidth]{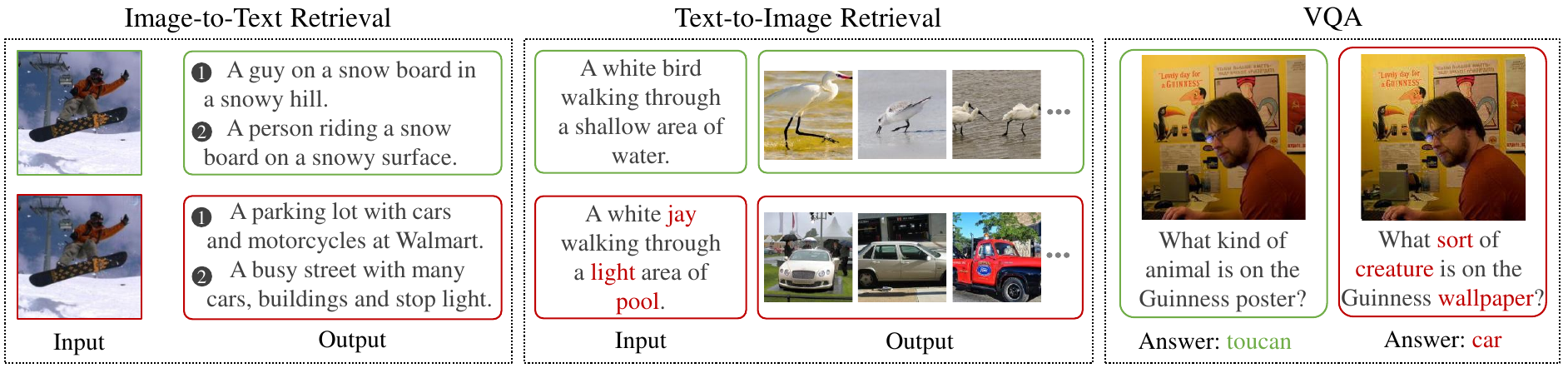}
    \end{center}
    \vspace{-2ex}
    \caption{\small Visualization of the attack results for different tasks. Green boxes indicate benign input samples and correct output outcomes, while red boxes represent backdoor samples and adversary-specified outcomes.}
    \label{fig:attack-visualization}
\end{figure*}

\noindent \textbf{Visualization} \cref{fig:modality-invariant-components} shows some examples of the modality-invariant factors extracted by the proposed cross-modal mining mechanism. From the visualization results, we can see that the conceived mechanism can effectively capture the mutual critical factors across the visual and linguistic modalities. Furthermore, we also provide real attack results for image-to-text retrieval, text-to-image retrieval, and VQA tasks in \cref{fig:attack-visualization} for intuitive understanding.

\section{Conclusion \label{sec:conclusion}}
In this paper, we first presented a new style of backdoor attacks, \textit{i.e.}, bilateral backdoors, to complement the scenarios of cross-modal attacks. We then developed a generalized invisible backdoor attack algorithm for cross-model learning, named \textit{BadCM}, which is the first attempt for diverse cross-modal attacks. Particularly, a novel cross-modal mining scheme was introduced to obtain modality-invariant components as the carrier of poisoning substances, together with the modality-specific trigger generators that can effectively conceal trigger patterns into restricted invariant regions of each modality.
Extensive experiments on cross-modal retrieval and VQA manifested that our method is superior to existing methods in generalization and stealthiness. Besides, further experiments against backdoor defenses  proved its resistance to defense.

{\small
\bibliographystyle{IEEEtran}
\normalem
\bibliography{egbib}

\begin{thebibliography}{10}
\providecommand{\url}[1]{#1}
\csname url@samestyle\endcsname
\providecommand{\newblock}{\relax}
\providecommand{\bibinfo}[2]{#2}
\providecommand{\BIBentrySTDinterwordspacing}{\spaceskip=0pt\relax}
\providecommand{\BIBentryALTinterwordstretchfactor}{4}
\providecommand{\BIBentryALTinterwordspacing}{\spaceskip=\fontdimen2\font plus
\BIBentryALTinterwordstretchfactor\fontdimen3\font minus
  \fontdimen4\font\relax}
\providecommand{\BIBforeignlanguage}[2]{{%
\expandafter\ifx\csname l@#1\endcsname\relax
\typeout{** WARNING: IEEEtran.bst: No hyphenation pattern has been}%
\typeout{** loaded for the language `#1'. Using the pattern for}%
\typeout{** the default language instead.}%
\else
\language=\csname l@#1\endcsname
\fi
#2}}
\providecommand{\BIBdecl}{\relax}
\BIBdecl

\bibitem{wu2023adversarial}
B.~Wu, L.~Liu, Z.~Zhu, Q.~Liu, Z.~He, and S.~Lyu, ``Adversarial machine
  learning: A systematic survey of backdoor attack, weight attack and
  adversarial example,'' \emph{arXiv preprint arXiv:2302.09457}, 2023.

\bibitem{szegedy2013intriguing}
C.~Szegedy, W.~Zaremba, I.~Sutskever, J.~Bruna, D.~Erhan, I.~Goodfellow, and
  R.~Fergus, ``Intriguing properties of neural networks,'' \emph{arXiv preprint
  arXiv:1312.6199}, 2013.

\bibitem{che2021adversarial}
Z.~Che, A.~Borji, G.~Zhai, S.~Ling, J.~Li, Y.~Tian, G.~Guo, and P.~Le~Callet,
  ``Adversarial attack against deep saliency models powered by non-redundant
  priors,'' \emph{IEEE Transactions on Image Processing}, vol.~30, pp.
  1973--1988, 2021.

\bibitem{gu2017badnets}
T.~Gu, B.~Dolan-Gavitt, and S.~Garg, ``Badnets: Identifying vulnerabilities in
  the machine learning model supply chain,'' \emph{arXiv preprint
  arXiv:1708.06733}, 2017.

\bibitem{chen2017targeted}
X.~Chen, C.~Liu, B.~Li, K.~Lu, and D.~Song, ``Targeted backdoor attacks on deep
  learning systems using data poisoning,'' \emph{arXiv preprint
  arXiv:1712.05526}, 2017.

\bibitem{zhang2022poison}
J.~Zhang, C.~Dongdong, Q.~Huang, J.~Liao, W.~Zhang, H.~Feng, G.~Hua, and N.~Yu,
  ``Poison ink: Robust and invisible backdoor attack,'' \emph{IEEE Transactions
  on Image Processing}, vol.~31, pp. 5691--5705, 2022.

\bibitem{feng2022fiba}
Y.~Feng, B.~Ma, J.~Zhang, S.~Zhao, Y.~Xia, and D.~Tao, ``Fiba:
  Frequency-injection based backdoor attack in medical image analysis,'' in
  \emph{Proceedings of the IEEE/CVF Conference on Computer Vision and Pattern
  Recognition}, 2022, pp. 20\,876--20\,885.

\bibitem{wang2022invisible}
T.~Wang, Y.~Yao, F.~Xu, S.~An, H.~Tong, and T.~Wang, ``An invisible black-box
  backdoor attack through frequency domain,'' in \emph{Computer Vision--ECCV
  2022: 17th European Conference, Tel Aviv, Israel, October 23--27, 2022,
  Proceedings, Part XIII}.\hskip 1em plus 0.5em minus 0.4em\relax Springer,
  2022, pp. 396--413.

\bibitem{chen2021badnl}
X.~Chen, A.~Salem, D.~Chen, M.~Backes, S.~Ma, Q.~Shen, Z.~Wu, and Y.~Zhang,
  ``Badnl: Backdoor attacks against nlp models with semantic-preserving
  improvements,'' in \emph{Annual Computer Security Applications Conference},
  2021, pp. 554--569.

\bibitem{qi2021hidden}
F.~Qi, M.~Li, Y.~Chen, Z.~Zhang, Z.~Liu, Y.~Wang, and M.~Sun, ``Hidden killer:
  Invisible textual backdoor attacks with syntactic trigger,'' in
  \emph{Proceedings of the 59th Annual Meeting of the Association for
  Computational Linguistics and the 11th International Joint Conference on
  Natural Language Processing}, 2021, pp. 443--453.

\bibitem{walmer2022dual}
M.~Walmer, K.~Sikka, I.~Sur, A.~Shrivastava, and S.~Jha, ``Dual-key multimodal
  backdoors for visual question answering,'' in \emph{Proceedings of the
  IEEE/CVF Conference on Computer Vision and Pattern Recognition}, 2022, pp.
  15\,375--15\,385.

\bibitem{wang2016comprehensive}
K.~Wang, Q.~Yin, W.~Wang, S.~Wu, and L.~Wang, ``A comprehensive survey on
  cross-modal retrieval,'' \emph{arXiv preprint arXiv:1607.06215}, 2016.

\bibitem{qin2022joint}
J.~Qin, L.~Fei, Z.~Zhang, J.~Wen, Y.~Xu, and D.~Zhang, ``Joint specifics and
  consistency hash learning for large-scale cross-modal retrieval,'' \emph{IEEE
  Transactions on Image Processing}, vol.~31, pp. 5343--5358, 2022.

\bibitem{antol2015vqa}
S.~Antol, A.~Agrawal, J.~Lu, M.~Mitchell, D.~Batra, C.~L. Zitnick, and
  D.~Parikh, ``Vqa: Visual question answering,'' in \emph{Proceedings of the
  IEEE international conference on computer vision}, 2015, pp. 2425--2433.

\bibitem{guo2021loss}
Y.~Guo, L.~Nie, Z.~Cheng, Q.~Tian, and M.~Zhang, ``Loss re-scaling vqa:
  revisiting the language prior problem from a class-imbalance view,''
  \emph{IEEE Transactions on Image Processing}, vol.~31, pp. 227--238, 2021.

\bibitem{xu2022multi}
L.~Xu, X.~Zeng, B.~Zheng, and W.~Li, ``Multi-manifold deep discriminative
  cross-modal hashing for medical image retrieval,'' \emph{IEEE Transactions on
  Image Processing}, vol.~31, pp. 3371--3385, 2022.

\bibitem{zhang2022deep}
Y.~Zhang, W.~Ou, Y.~Shi, J.~Deng, X.~You, and A.~Wang, ``Deep medical
  cross-modal attention hashing,'' \emph{World Wide Web}, vol.~25, no.~4, pp.
  1519--1536, 2022.

\bibitem{gurari2018vizwiz}
D.~Gurari, Q.~Li, A.~J. Stangl, A.~Guo, C.~Lin, K.~Grauman, J.~Luo, and J.~P.
  Bigham, ``Vizwiz grand challenge: Answering visual questions from blind
  people,'' in \emph{Proceedings of the IEEE conference on computer vision and
  pattern recognition}, 2018, pp. 3608--3617.

\bibitem{li2022object}
M.~Li, N.~Zhong, X.~Zhang, Z.~Qian, and S.~Li, ``Object-oriented backdoor
  attack against image captioning,'' in \emph{ICASSP 2022-2022 IEEE
  International Conference on Acoustics, Speech and Signal Processing
  (ICASSP)}.\hskip 1em plus 0.5em minus 0.4em\relax IEEE, 2022, pp. 2864--2868.

\bibitem{liu2018fine}
K.~Liu, B.~Dolan-Gavitt, and S.~Garg, ``Fine-pruning: Defending against
  backdooring attacks on deep neural networks,'' in \emph{Research in Attacks,
  Intrusions, and Defenses: 21st International Symposium, RAID 2018, Heraklion,
  Crete, Greece, September 10-12, 2018, Proceedings 21}.\hskip 1em plus 0.5em
  minus 0.4em\relax Springer, 2018, pp. 273--294.

\bibitem{doan2020februus}
B.~G. Doan, E.~Abbasnejad, and D.~C. Ranasinghe, ``Februus: Input purification
  defense against trojan attacks on deep neural network systems,'' in
  \emph{Annual Computer Security Applications Conference}, 2020, pp. 897--912.

\bibitem{chen2022effective}
W.~Chen, B.~Wu, and H.~Wang, ``Effective backdoor defense by exploiting
  sensitivity of poisoned samples,'' \emph{Advances in Neural Information
  Processing Systems}, vol.~35, pp. 9727--9737, 2022.

\bibitem{anderson2018bottom}
P.~Anderson, X.~He, C.~Buehler, D.~Teney, M.~Johnson, S.~Gould, and L.~Zhang,
  ``Bottom-up and top-down attention for image captioning and visual question
  answering,'' in \emph{Proceedings of the IEEE conference on computer vision
  and pattern recognition}, 2018, pp. 6077--6086.

\bibitem{lee2018stacked}
K.-H. Lee, X.~Chen, G.~Hua, H.~Hu, and X.~He, ``Stacked cross attention for
  image-text matching,'' in \emph{Proceedings of the European conference on
  computer vision (ECCV)}, 2018, pp. 201--216.

\bibitem{nguyen2020wanet}
T.~A. Nguyen and A.~T. Tran, ``Wanet-imperceptible warping-based backdoor
  attack,'' in \emph{International Conference on Learning Representations},
  2020.

\bibitem{lin2020composite}
J.~Lin, L.~Xu, Y.~Liu, and X.~Zhang, ``Composite backdoor attack for deep
  neural network by mixing existing benign features,'' in \emph{Proceedings of
  the 2020 ACM SIGSAC Conference on Computer and Communications Security},
  2020, pp. 113--131.

\bibitem{nguyen2020input}
T.~A. Nguyen and A.~Tran, ``Input-aware dynamic backdoor attack,''
  \emph{Advances in Neural Information Processing Systems}, vol.~33, pp.
  3454--3464, 2020.

\bibitem{li2021invisible}
Y.~Li, Y.~Li, B.~Wu, L.~Li, R.~He, and S.~Lyu, ``Invisible backdoor attack with
  sample-specific triggers,'' in \emph{Proceedings of the IEEE/CVF
  International Conference on Computer Vision}, 2021, pp. 16\,463--16\,472.

\bibitem{jiang2023color}
W.~Jiang, H.~Li, G.~Xu, and T.~Zhang, ``Color backdoor: A robust poisoning
  attack in color space,'' in \emph{Proceedings of the IEEE/CVF Conference on
  Computer Vision and Pattern Recognition}, 2023, pp. 8133--8142.

\bibitem{doan2021lira}
K.~Doan, Y.~Lao, W.~Zhao, and P.~Li, ``Lira: Learnable, imperceptible and
  robust backdoor attacks,'' in \emph{Proceedings of the IEEE/CVF International
  Conference on Computer Vision}, 2021, pp. 11\,966--11\,976.

\bibitem{zhao2022defeat}
Z.~Zhao, X.~Chen, Y.~Xuan, Y.~Dong, D.~Wang, and K.~Liang, ``Defeat: Deep
  hidden feature backdoor attacks by imperceptible perturbation and latent
  representation constraints,'' in \emph{Proceedings of the IEEE/CVF Conference
  on Computer Vision and Pattern Recognition}, 2022, pp. 15\,213--15\,222.

\bibitem{dai2019backdoor}
J.~Dai, C.~Chen, and Y.~Li, ``A backdoor attack against lstm-based text
  classification systems,'' \emph{IEEE Access}, vol.~7, pp. 138\,872--138\,878,
  2019.

\bibitem{kurita2020weight}
K.~Kurita, P.~Michel, and G.~Neubig, ``Weight poisoning attacks on pretrained
  models,'' in \emph{Proceedings of the 58th Annual Meeting of the Association
  for Computational Linguistics}, 2020, pp. 2793--2806.

\bibitem{chen2022kallima}
X.~Chen, Y.~Dong, Z.~Sun, S.~Zhai, Q.~Shen, and Z.~Wu, ``Kallima: A clean-label
  framework for textual backdoor attacks,'' in \emph{Computer Security--ESORICS
  2022: 27th European Symposium on Research in Computer Security, Copenhagen,
  Denmark, September 26--30, 2022, Proceedings, Part I}.\hskip 1em plus 0.5em
  minus 0.4em\relax Springer, 2022, pp. 447--466.

\bibitem{qi2021turn}
F.~Qi, Y.~Yao, S.~Xu, Z.~Liu, and M.~Sun, ``Turn the combination lock:
  Learnable textual backdoor attacks via word substitution,'' in
  \emph{Proceedings of the 59th Annual Meeting of the Association for
  Computational Linguistics and the 11th International Joint Conference on
  Natural Language Processing}, 2021, pp. 4873--4883.

\bibitem{gan2022triggerless}
L.~Gan, J.~Li, T.~Zhang, X.~Li, Y.~Meng, F.~Wu, Y.~Yang, S.~Guo, and C.~Fan,
  ``Triggerless backdoor attack for nlp tasks with clean labels,'' in
  \emph{Proceedings of the 2022 Conference of the North American Chapter of the
  Association for Computational Linguistics: Human Language Technologies},
  2022, pp. 2942--2952.

\bibitem{jiang2017deep}
Q.-Y. Jiang and W.-J. Li, ``Deep cross-modal hashing,'' in \emph{Proceedings of
  the IEEE conference on computer vision and pattern recognition}, 2017, pp.
  3232--3240.

\bibitem{wang2017adversarial}
B.~Wang, Y.~Yang, X.~Xu, A.~Hanjalic, and H.~T. Shen, ``Adversarial cross-modal
  retrieval,'' in \emph{Proceedings of the 25th ACM international conference on
  Multimedia}, 2017, pp. 154--162.

\bibitem{zhen2019deep}
L.~Zhen, P.~Hu, X.~Wang, and D.~Peng, ``Deep supervised cross-modal
  retrieval,'' in \emph{Proceedings of the IEEE/CVF Conference on Computer
  Vision and Pattern Recognition}, 2019, pp. 10\,394--10\,403.

\bibitem{xu2015show}
K.~Xu, J.~Ba, R.~Kiros, K.~Cho, A.~Courville, R.~Salakhudinov, R.~Zemel, and
  Y.~Bengio, ``Show, attend and tell: Neural image caption generation with
  visual attention,'' in \emph{International conference on machine
  learning}.\hskip 1em plus 0.5em minus 0.4em\relax PMLR, 2015, pp. 2048--2057.

\bibitem{zhou2019re}
L.~Zhou, Y.~Zhang, Y.-G. Jiang, T.~Zhang, and W.~Fan, ``Re-caption:
  Saliency-enhanced image captioning through two-phase learning,'' \emph{IEEE
  Transactions on Image Processing}, vol.~29, pp. 694--709, 2019.

\bibitem{yu2019deep}
Z.~Yu, J.~Yu, Y.~Cui, D.~Tao, and Q.~Tian, ``Deep modular co-attention networks
  for visual question answering,'' in \emph{Proceedings of the IEEE/CVF
  conference on computer vision and pattern recognition}, 2019, pp. 6281--6290.

\bibitem{saha2020hidden}
A.~Saha, A.~Subramanya, and H.~Pirsiavash, ``Hidden trigger backdoor attacks,''
  in \emph{Proceedings of the AAAI conference on artificial intelligence},
  vol.~34, no.~07, 2020, pp. 11\,957--11\,965.

\bibitem{ren2015faster}
S.~Ren, K.~He, R.~Girshick, and J.~Sun, ``Faster r-cnn: Towards real-time
  object detection with region proposal networks,'' \emph{Advances in neural
  information processing systems}, vol.~28, 2015.

\bibitem{radford2021learning}
A.~Radford, J.~W. Kim, C.~Hallacy, A.~Ramesh, G.~Goh, S.~Agarwal, G.~Sastry,
  A.~Askell, P.~Mishkin, J.~Clark \emph{et~al.}, ``Learning transferable visual
  models from natural language supervision,'' in \emph{International Conference
  on Machine Learning}.\hskip 1em plus 0.5em minus 0.4em\relax PMLR, 2021, pp.
  8748--8763.

\bibitem{isola2017image}
P.~Isola, J.-Y. Zhu, T.~Zhou, and A.~A. Efros, ``Image-to-image translation
  with conditional adversarial networks,'' in \emph{Proceedings of the IEEE
  conference on computer vision and pattern recognition}, 2017, pp. 1125--1134.

\bibitem{li2020bert}
L.~Li, R.~Ma, Q.~Guo, X.~Xue, and X.~Qiu, ``Bert-attack: Adversarial attack
  against bert using bert,'' \emph{arXiv preprint arXiv:2004.09984}, 2020.

\bibitem{devlin2018bert}
J.~Devlin, M.-W. Chang, K.~Lee, and K.~Toutanova, ``Bert: Pre-training of deep
  bidirectional transformers for language understanding,'' \emph{arXiv preprint
  arXiv:1810.04805}, 2018.

\bibitem{chua2009nus}
T.~Chua, J.~Tang, R.~Hong, H.~Li, Z.~Luo, and Y.~Zheng, ``Nus-wide: A
  real-world web image database from national university of singapore,'' in
  \emph{ACM International Conference on Image and Video Retrieval}, 2009, pp.
  1--9.

\bibitem{lin2014microsoft}
T.~Lin, M.~Maire, S.~Belongie, J.~Hays, P.~Perona, D.~Ramanan, P.~Doll{\'a}r,
  and C.~L. Zitnick, ``Microsoft coco: Common objects in context,'' in
  \emph{European Conference on Computer Vision}, 2014, pp. 740--755.

\bibitem{escalante2010segmented}
H.~J. Escalante, C.~A. Hern{\'a}ndez, J.~A. Gonzalez, A.~L{\'o}pez-L{\'o}pez,
  M.~Montes, E.~F. Morales, L.~E. Sucar, L.~Villasenor, and M.~Grubinger, ``The
  segmented and annotated iapr tc-12 benchmark,'' \emph{Computer vision and
  image understanding}, vol. 114, no.~4, pp. 419--428, 2010.

\bibitem{gao2021clean}
K.~Gao, J.~Bai, B.~Chen, D.~Wu, and S.-T. Xia, ``Clean-label backdoor attack
  against deep hashing based retrieval,'' \emph{arXiv preprint
  arXiv:2109.08868}, 2021.

\bibitem{huynh2008scope}
Q.~Huynh-Thu and M.~Ghanbari, ``Scope of validity of psnr in image/video
  quality assessment,'' \emph{Electronics letters}, vol.~44, no.~13, pp.
  800--801, 2008.

\bibitem{ronneberger2015u}
O.~Ronneberger, P.~Fischer, and T.~Brox, ``U-net: Convolutional networks for
  biomedical image segmentation,'' in \emph{International Conference on Medical
  image computing and computer-assisted intervention}.\hskip 1em plus 0.5em
  minus 0.4em\relax Springer, 2015, pp. 234--241.

\bibitem{qi2021mind}
F.~Qi, Y.~Chen, X.~Zhang, M.~Li, Z.~Liu, and M.~Sun, ``Mind the style of text!
  adversarial and backdoor attacks based on text style transfer,'' in
  \emph{Proceedings of the 2021 Conference on Empirical Methods in Natural
  Language Processing}, 2021, pp. 4569--4580.

\bibitem{hayase2021spectre}
J.~Hayase, W.~Kong, R.~Somani, and S.~Oh, ``Spectre: Defending against backdoor
  attacks using robust statistics,'' in \emph{International Conference on
  Machine Learning}.\hskip 1em plus 0.5em minus 0.4em\relax PMLR, 2021, pp.
  4129--4139.

\bibitem{selvaraju2017grad}
R.~R. Selvaraju, M.~Cogswell, A.~Das, R.~Vedantam, D.~Parikh, and D.~Batra,
  ``Grad-cam: Visual explanations from deep networks via gradient-based
  localization,'' in \emph{Proceedings of the IEEE international conference on
  computer vision}, 2017, pp. 618--626.

\bibitem{qi2020onion}
F.~Qi, Y.~Chen, M.~Li, Y.~Yao, Z.~Liu, and M.~Sun, ``Onion: A simple and
  effective defense against textual backdoor attacks,'' \emph{arXiv preprint
  arXiv:2011.10369}, 2020.

\bibitem{kim2021vilt}
W.~Kim, B.~Son, and I.~Kim, ``Vilt: Vision-and-language transformer without
  convolution or region supervision,'' in \emph{International Conference on
  Machine Learning}.\hskip 1em plus 0.5em minus 0.4em\relax PMLR, 2021, pp.
  5583--5594.

\bibitem{simonyan2014very}
K.~Simonyan and A.~Zisserman, ``Very deep convolutional networks for
  large-scale image recognition,'' \emph{arXiv preprint arXiv:1409.1556}, 2014.

\bibitem{he2016deep}
K.~He, X.~Zhang, S.~Ren, and J.~Sun, ``Deep residual learning for image
  recognition,'' in \emph{Proceedings of the IEEE conference on computer vision
  and pattern recognition}, 2016, pp. 770--778.

\bibitem{kim2018bilinear}
J.-H. Kim, J.~Jun, and B.-T. Zhang, ``Bilinear attention networks,''
  \emph{Advances in neural information processing systems}, vol.~31, 2018.

\bibitem{yu2017multi}
Z.~Yu, J.~Yu, J.~Fan, and D.~Tao, ``Multi-modal factorized bilinear pooling
  with co-attention learning for visual question answering,'' in
  \emph{Proceedings of the IEEE international conference on computer vision},
  2017, pp. 1821--1830.

\bibitem{yu2018beyond}
Z.~Yu, J.~Yu, C.~Xiang, J.~Fan, and D.~Tao, ``Beyond bilinear: Generalized
  multimodal factorized high-order pooling for visual question answering,''
  \emph{IEEE transactions on neural networks and learning systems}, vol.~29,
  no.~12, pp. 5947--5959, 2018.

\bibitem{goyal2017making}
Y.~Goyal, T.~Khot, D.~Summers-Stay, D.~Batra, and D.~Parikh, ``Making the v in
  vqa matter: Elevating the role of image understanding in visual question
  answering,'' in \emph{Proceedings of the IEEE conference on computer vision
  and pattern recognition}, 2017, pp. 6904--6913.

\end{thebibliography}
}

\begin{IEEEbiography}[{\includegraphics[width=1in,height=1.25in,clip,keepaspectratio]{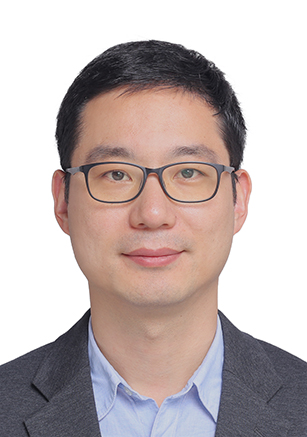}}]{Zheng Zhang} (Senior Member, IEEE) received the Ph.D. degree from the Harbin Institute of Technology, China. He was a Postdoctoral Research Fellow at The University of Queensland, Australia. He is currently with the Harbin Institute of Technology, Shenzhen, China. He has co-authored more than 100 technical papers in prestigious international journals and conferences. His research interests include multimedia content analysis and understanding. He is an Associate Editor of  IEEE TAFFC, IEEE JBHI, and others and also serves as an Area Chair of ICML, CVPR, ACM MM, etc.
\end{IEEEbiography}

\begin{IEEEbiography}[{\includegraphics[width=1in,height=1.25in,clip,keepaspectratio]{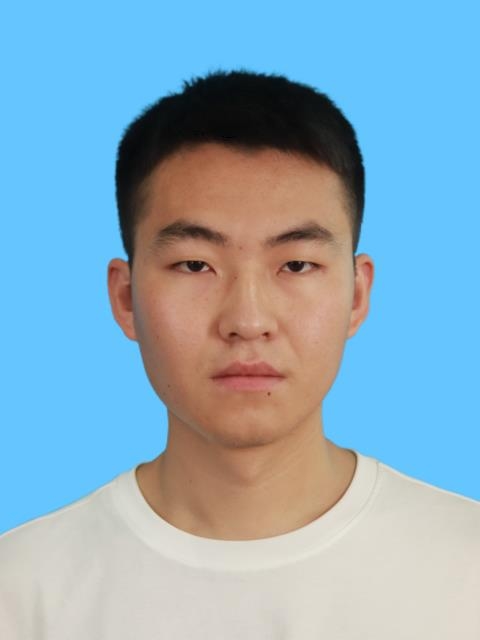}}]{Xu Yuan} 
received the B.S. degree from the Harbin Institute of Technology, Weihai, China, in 2021. He is currently pursuing the master’s degree with the Harbin Institute of Technology, Shenzhen, China. His research interests include deep learning and adversarial machine learning.
\end{IEEEbiography}

\begin{IEEEbiography}[{\includegraphics[width=1in,height=1.25in,clip,keepaspectratio]{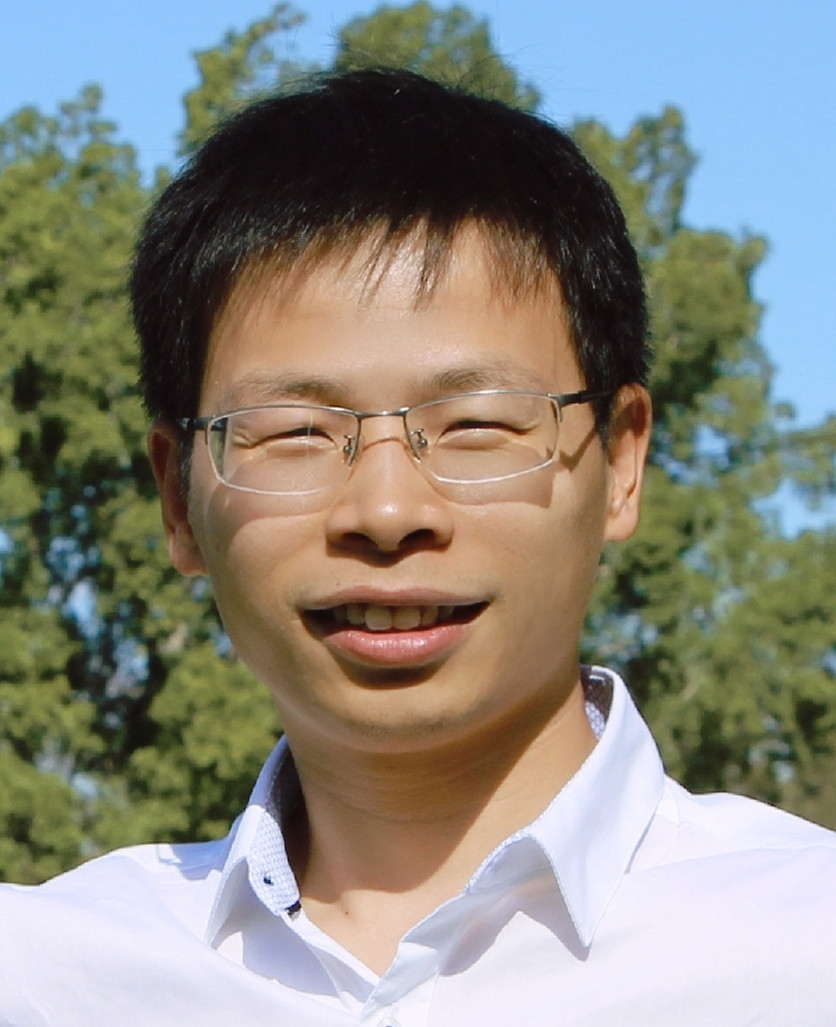}}]{Lei Zhu} (Senior Member, IEEE) 
is currently with School of Electronic and Information Engineering, Tongji University, China. He received his B.Eng. and Ph.D. degrees from Wuhan University of Technology in 2009 and Huazhong University Science and Technology in 2015, respectively. He was a Postdoctoral Research Fellow at the University of Queensland (2016-2017). His research interests are in the area of large-scale multimedia content analysis and retrieval. Zhu has co-/authored more than 100 peer-reviewed papers with more than 7,200 Google citations. He serves as the Associate Editor of IEEE TBD and ACM TOMM. He has served as the Area Chair for ACM MM and IEEE ICME, Senior Program Committee for SIGIR, AAAI, and CIKM.
\end{IEEEbiography}

\begin{IEEEbiography}[{\includegraphics[width=1in,height=1.25in,clip,keepaspectratio]{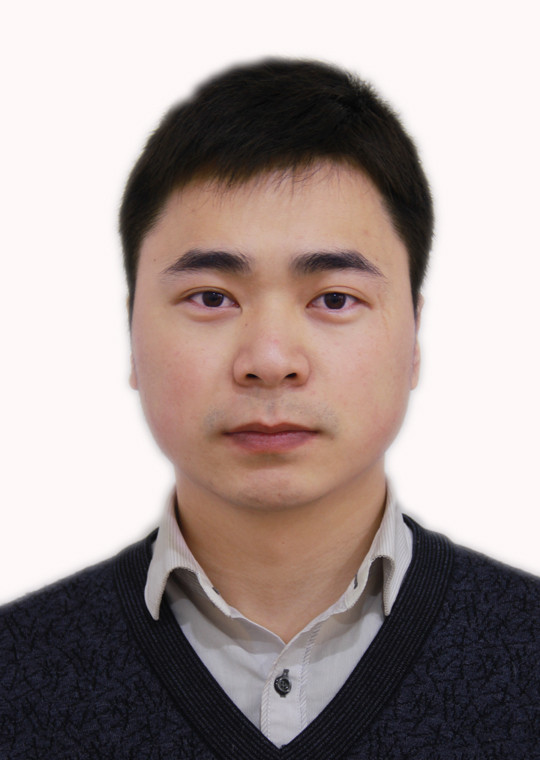}}]{Jingkuan Song}
is a full professor with University of Electronic Science and Technology of China (UESTC). He joined Columbia University as a Postdoc Research Scientist (2016-2017), and University of Trento as a Research Fellow (2014-2016). He obtained his PhD degree from The University of Queensland (UQ), Australia. His research interests mainly focus on Multimedia Compact Repersentation and Analysis. He was the winner of the Best Paper Award in ICPR (2016, Mexico), Best Student Paper Award in Australian Database Conference (2017, Australia), Best Paper Honorable Mention Award in SIGIR (2017, Japan), Best Paper Runner-up Award in ApWEB (2019, China) and ACM SIGMM Rising Star Award 2021. He is Associate Editor of IEEE TMM, ACM TOMM, Guest Editor of TMM, PR and AC/SPC/PC member of CVPR’18-'23, ACM MM'18-'23, IJCAI'18-'23, etc.
\end{IEEEbiography}

\begin{IEEEbiography}[{\includegraphics[width=1in,height=1.25in,clip,keepaspectratio]{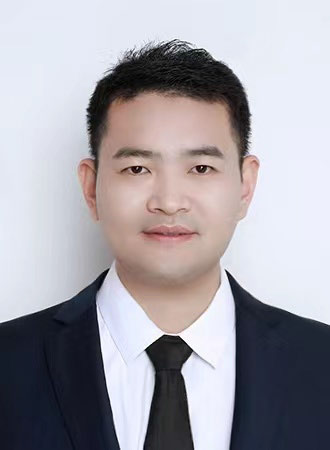}}]{Liqiang Nie} (Senior Member, IEEE) received the B.Eng. degree from Xi’an Jiaotong University and the Ph.D. degree from the National University of Singapore (NUS). After the Ph.D. degree, he continued his research with NUS, as a Research Fellow for three years. He is currently a Professor with the Harbin Institute of Technology (Shenzhen). He has coauthored more than 200 articles and four books. He received more than 19,000 Google Scholar citations as of May 2023. His research interests include multimedia computing and information retrieval. He received many awards, such as ACM MM and SIGIR Best Paper Honorable Mention in 2019, the SIGMM Rising Star in 2020, the TR35 China 2020, the DAMO Academy Young Fellow in 2020, the SIGIR Best Student Paper in 2021, and ACM MM Best Paper in 2022. He is the Area Chair of ACM MM from 2018 to 2023. He is an Associate Editor of IEEE TRANSACTIONS ON KNOWLEDGE AND DATA ENGINEERING, IEEE TRANSACTIONS ON MULTIMEDIA, ACM Transactions on Multimedia Computing, Communications, and Applications, and Information Sciences.
\end{IEEEbiography}

\vfill

\end{document}